\documentclass{article}

\usepackage[preprint]{neurips_2025}
\usepackage[utf8]{inputenc} 
\usepackage[T1]{fontenc}    
\usepackage{hyperref}       
\usepackage{url}            
\usepackage{booktabs}       
\usepackage{amsfonts}       
\usepackage{nicefrac}       
\usepackage{microtype}      
\usepackage{xcolor}         
\usepackage{amsmath}
\usepackage{graphicx}
\usepackage{amssymb}
\usepackage{float} 
\usepackage{multirow}
\usepackage{colortbl}
\usepackage{caption}

\newcommand{\model}{PEVA}
\newcommand{\ours}{PEVA}

\title{Whole-Body Conditioned \\ Egocentric Video Prediction}
\author{%
  Yutong Bai$^{*\:1}$  \hspace{.4in} Danny Tran$^{*\: 1}$ \hspace{.4in}  Amir Bar$^{*\: 2}$ \\ \\
     \textbf{Yann LeCun}$^{\dagger\:2,3}$  \hspace{.2in}  \textbf{Trevor Darrell}$^{\dagger\: 1}$\hspace{.2in} \textbf{Jitendra Malik}$^{\dagger\: 1,2}$ \\ \\
    $^1$UC Berkeley (BAIR) \hspace{.5in} $^2$FAIR, Meta \hspace{.5in} $^3$New York University
}

\begin{document}
\maketitle
\newenvironment{alphafootnotes}
  {\par\edef\savedfootnotenumber{\number\value{footnote}}
   \renewcommand{\thefootnote}{\alph{footnote}}
   \setcounter{footnote}{0}}
  {\par\setcounter{footnote}{\savedfootnotenumber}}
\begin{alphafootnotes}
\phantomsection\let\thefootnote\relax\footnotetext{* Equal contribution; $\dagger$ Equal advising.}
\end{alphafootnotes}

\begin{abstract}
We train models to \textbf{P}redict \textbf{E}go-centric \textbf{V}ideo from human \textbf{A}ctions (\textbf{\ours}), given the past video and an action represented by the relative 3D body pose. By conditioning on kinematic pose trajectories, structured by the joint hierarchy of the body, our model learns to simulate how physical human actions shape the environment from a first-person point of view. We train an auto-regressive conditional diffusion transformer on Nymeria, a large-scale dataset of real-world egocentric video and body pose capture. We further design a hierarchical evaluation protocol with increasingly challenging tasks, enabling a comprehensive analysis of the model’s embodied prediction and control abilities. Our work represents an initial attempt to tackle the challenges of modeling complex real-world environments and embodied agent behaviors with video prediction from the perspective of a human.\footnote{Project page:~\url{https://dannytran123.github.io/PEVA}.}
\end{abstract}

\section{Introduction}
Human movement is rich, continuous, and physically grounded~\citep{rosenhahn2008human,aggarwal1999human}. The way we walk, lean, turn, or reach—often subtle and coordinated—directly shapes what we see from a first-person perspective. For embodied agents to simulate and plan like humans, they must not only predict future observations~\citep{von1925helmholtz}, but also understand how visual input arises from whole-body action~\citep{craik1943nature}. This understanding is essential because many aspects of the environment are not immediately visible--we need to move our bodies to reveal new information and achieve our goals.

Vision serves as a natural signal for long-term planning~\citep{lecun2022path,hafner2023mastering, ebert2018visual,ma2022vip}. We look at our environment to plan and act, using our egocentric view as a predictive goal~\citep{sridhar2024nomad,bar2024navigation}. 
When we consider our body movements, we should consider both actions of the feet (locomotion and navigation) and the actions of the hand (manipulation), or more generally, whole-body control~\citep{nvidia2025gr00t,cheng2024expressive,he2024learning,radosavovic2024learning,he2024omnih2o,hansen2024hierarchical}.
For example, when reaching for an object, we must anticipate how our arm movement will affect what we see, even before the object comes into view. This ability to plan based on partial visual information is crucial for embodied agents to operate effectively in real-world environments.

\begin{figure*}
    \centering
    \includegraphics[width=1\linewidth]{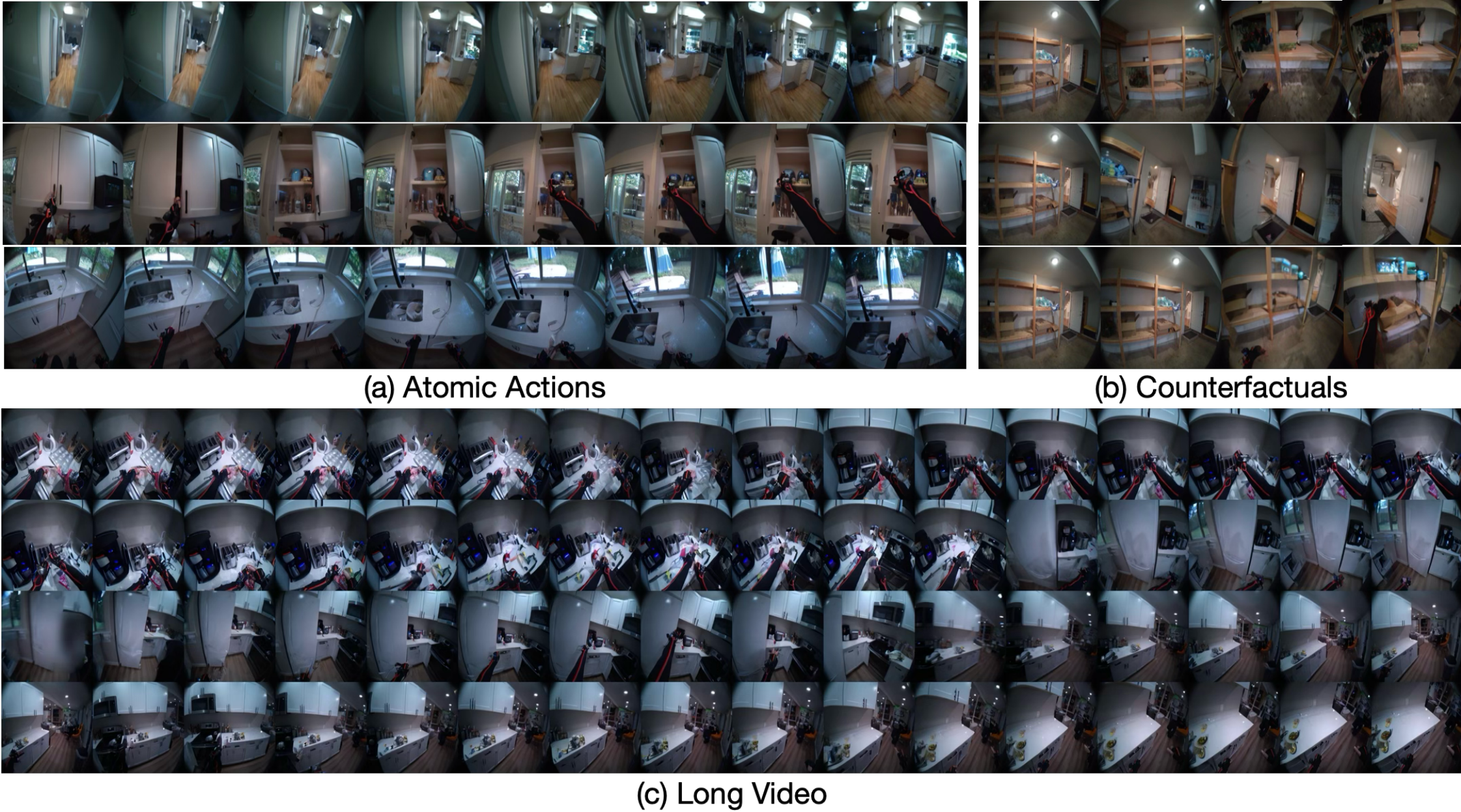}
    \vspace{-0.5cm}
\caption{\textbf{Predicting Ego-centric Video from human Actions (PEVA)}. Given past video frames and an action specifying a desired change in 3D pose, \ours~predicts the next video frame. Our results show that, given the first frame and a sequence of actions, our model can generate videos of atomic actions (a), simulate counterfactuals (b), and support long video generation (c).}
    \label{fig:teaser}
    \vspace{-0.7cm}
\end{figure*}

Building a model that can effectively learn from and predict based on whole-body motion presents several fundamental challenges. First, representing human actions requires capturing both global body dynamics and fine-grained joint articulations, which involves high-dimensional, structured data with complex temporal dependencies. Second, the relationship between body movements and visual perception is highly nonlinear and context-dependent—the same arm movement can result in different visual outcomes depending on the environment and the agent's current state. Third, learning these relationships from real-world data is particularly challenging due to the inherent variability in human motion and the subtle, often delayed visual consequences of actions.

To address these challenges, we develop a novel approach \textbf{\ours} that combines several key innovations. First, we design a structured action representation that preserves both global body dynamics and local joint movements, using a hierarchical encoding that captures the kinematic tree structure of human motion. This representation enables the model to understand both the overall body movement and the fine-grained control of individual joints. Second, we develop a novel architecture based on conditional diffusion transformers that can effectively model the complex, nonlinear relationship between body movements and visual outcomes. The architecture incorporates temporal attention mechanisms to capture long-range dependencies and a specialized action embedding component that maintains the structured nature of human motion. Third, we leverage a large-scale dataset of synchronized egocentric video and motion capture data~\citep{ma2024nymeria}, which provides the necessary training signal to learn these complex relationships. Our training strategy includes random timeskips to handle the delayed visual consequences of actions and sequence-level training to maintain temporal coherence.

For evaluation, we first assess the model's ability to predict immediate visual consequences by evaluating its performance on single-step predictions at 2-second intervals, measuring both perceptual quality (LPIPS~\citep{zhang2018unreasonable}) and semantic consistency (DreamSim~\citep{fu2023dreamsim}). Second, we decompose complex human movements into atomic actions—such as hand movements (up, down, left, right) and whole-body movements (forward, rotation)—to test the model's understanding of how specific joint-level movements affect the egocentric view. This fine-grained analysis reveals whether the model can capture the nuanced relationship between individual joint movements and their visual effects. Third, we examine the model's capability to predict long-term visual consequences by evaluating its performance across extended time horizons (up to 16 seconds), where the effects of actions may be delayed or not immediately visible.  Finally, we explore the model's ability to serve as a world model for planning by using it to simulate actions and choose the ones that lead to a predefined goal. This layered approach allows us to systematically analyze the strengths and limitations of our model, revealing both its capacity to simulate embodied perception and the open challenges that remain in bridging the gap between physical action and visual experience.

To conclude, we introduce \model{}, a model that predicts future egocentric video conditioned on whole-body human motion. By leveraging structured action representations derived from 3D pose trajectories, our model captures the intricate relationship between physical movement and visual perception. We develop a diffusion-based architecture that can be trained auto-regressively over a sequence of images in a parallelized fashion. Our model utilizes random time-skips that enable covering long-term videos efficiently. Trained on Nymeria~\citep{ma2024nymeria}, a large-scale real-world dataset of egocentric video and synchronized motion, \model{} advances embodied simulation with physically grounded, visually realistic predictions. Our comprehensive evaluation framework demonstrates that whole-body control significantly improves video quality, semantic consistency, and simulating counterfactual.

\section{Related Works}

\noindent \textbf{World Models.}
The concept of a ``world model'', an internal representation of the world used for prediction and planning, has a rich history across multiple disciplines. The idea was first proposed in psychology by \citet{craik1943nature}, who hypothesized that the brain uses ``small-scale models'' of reality to anticipate events. This principle found parallel development in control theory, where methods like the Kalman Filter and Linear Quadratic Regulator~(LQR) rely on an explicit model of the system to be controlled~\citep{kalman1960new}. The idea of internal models became central to computational neuroscience for explaining motor control, with researchers proposing that the brain plans and executes movements by simulating them first~\citep{jordan1996computational,kawato1987hierarchical,kawato1999internal}.

With the rise of deep learning, the focus shifted to learning these predictive models directly from data. Early work in computer vision demonstrated that models could learn intuitive physics from visual data to solve simple control tasks like playing billiards or poking objects~\citep{fragkiadaki2015learning,agrawal2016learning}. This paved the way for modern, large-scale world models that predict future video frames conditioned on actions, enabling planning by ``imagining'' future outcomes~\citep{ha2018world,hafnerdream,liu2024multi,li20223d,zhou2024dino,yang2023learning,yang2024video,assran2025v}. In reinforcement learning, models like Dreamer have shown that learning a world model improves sample efficiency~\citep{hafner2023mastering}. Recent approaches have used diffusion models for more expressive generation; for example, DIAMOND generates multi-step rollouts via autoregressive diffusion~\citep{alonso2024diamond}. In the egocentric domain, Navigation World Models (NWM) used conditional diffusion transformers (CDiT) to predict future frames from a planned trajectory~\citep{bar2024navigation}. However, these models use low-dimensional controls and neglect the agent's own body dynamics. We build on this extensive line of work by conditioning video prediction on whole-body pose, enabling a more physically-grounded simulation.

\noindent \textbf{Human Motion Generation and Controllable Prediction.}
Human motion modeling has advanced from recurrent and VAE-based methods~\citep{rempe2021humor,petrovich2021action,ye2023decoupling} to powerful diffusion-based generators~\citep{tevet2022human,zhang2024motiondiffuse}. These models generate diverse, realistic 3D pose sequences conditioned on text~\citep{hong2024egolm,guo2022generating,dabral2023mofusion}, audio~\citep{ng2024audio,dabral2023mofusion,ao2023gesturediffuclip}, and head pose~\citep{li2023egoego,castillo2023bodiffusion,yi2025egoallo}. Recent works like Animate Anyone~\citep{hu2023animate} and MagicAnimate~\citep{xu2023magicanimate} generate high-fidelity human animations from a reference image and pose sequence. Physically-aware extensions like PhysDiff~\citep{yuan2023physdiff} incorporate contact into the denoising loop. While prior works treat pose as the target, our model uses it as input for egocentric video prediction, reversing the typical motion generation setup. This enables fine-grained visual control, bridging pose-conditioned video generation~\citep{wu2023tune,zhang2023adding} with embodied simulation. Unlike Make-a-Video~\citep{singer2022make} or Tune-A-Video~\citep{wu2023tune}, which focus on text/image prompts, we condition directly on physically realizable body motion.

\noindent \textbf{Egocentric Perception and Embodied Forecasting.} Egocentric video datasets such as Ego4D~\citep{grauman2022ego4d}, Ego-Exo4D~\citep{grauman2024ego} and EPIC-KITCHENS~\citep{damen2018scaling} were used to study human action recognition, object anticipation~\citep{furnari2020rolling}, future video prediction~\citep{girdhar2021anticipative}, and even animal behavior~\citep{bar2024egopet}. To study pose estimation, EgoBody~\citep{zhang2022egobody} and Nymeria~\citep{ma2024nymeria} provide synchronized egocentric video and 3D pose. Unlike these works, we treat future body motion as a control signal, enabling visually grounded rollout. Prior works in egocentric pose forecasting~\citep{yuan2019ego} and visual foresight~\citep{finn2017deep} show that predicting future perception supports downstream planning. Our model unifies these lines by predicting future egocentric video from detailed whole-body control, enabling first-person planning with physical and visual realism.

\section{PEVA}
\label{sec:method}
In this section we describe our whole-body-conditioned ego-centric video prediction model. We start by describing how to represent human actions (Section~\ref{sec:actions}), then move on to describe the model and the training objective (Section~\ref{sec:obj}). Finally, we describe the model architecture in Section~\ref{sec:arch}.

\begin{figure}
    \centering
    \includegraphics[width=1\linewidth]{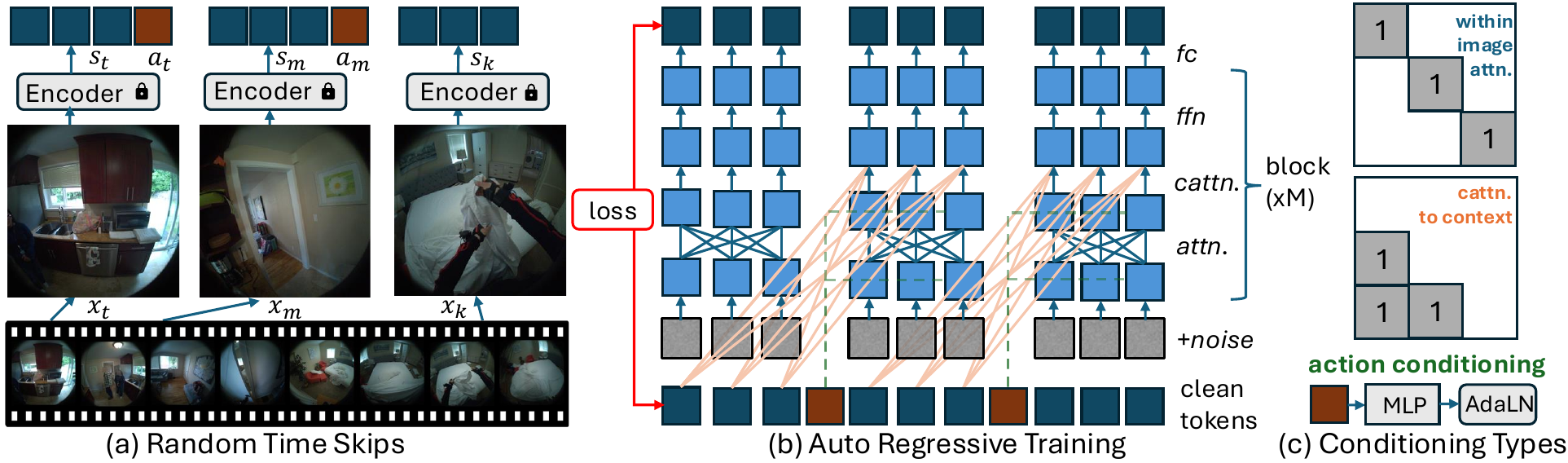}
    \caption{\textbf{Design of~\ours.} To train on an input video, we choose a random subset of frames and encode them via a fixed encoder (a). They are then fed to a CDiT that is trained autoregressively with teacher forcing (b). During the denoising process, each token attends to same-image tokens and cross-attends to clean tokens from past image(s). Action conditioning is done via AdaLN layers.}
    \vspace{-0.4cm}
    \label{fig:arch}
\end{figure}

\subsection{Structured Action Representation from Motion Data}
\label{sec:actions}
To effectively capture the relationship between human motion and egocentric visual perception, we define each action as a high-dimensional vector encoding both global body dynamics and detailed joint articulations. Rather than relying on simplified or discrete controls, our framework uses full-body motion information, including global translation (via the root joint) and relative joint rotations structured by the kinematic tree. This design ensures that the action space richly represents human movement at both coarse and fine levels.

To construct this representation, we synchronize motion capture data with video frames based on timestamps, then transform global coordinates into a local frame centered at the pelvis. This transformation makes the data invariant to initial position and orientation. Global positions are converted to local coordinates, quaternions to relative Euler angles, and joint relationships are preserved using the kinematic hierarchy. We normalize all motion parameters for stable learning: positions are scaled to $[-1, 1]$ and rotations bounded within $[-\pi, \pi]$. Each action reflects the change between consecutive frames, capturing motion over time and enabling the model to learn how physical actions translate into visual outcomes.

\subsection{Ego-Centric Video Prediction for Whole-Body Control}
\label{sec:obj}

Next, we describe our formulation of \textbf{\ours} from the perspective of an embodied agent. Intuitively, the model is an autoregressive diffusion model that receives an input video and a corresponding sequence of actions describing how the agent moves and acts. Given any prefix of frames and actions, the model predicts the resulting state of the world after applying the last action and considering other environment dynamics.  

More formally, we are given a dataset $D = \{(x_0, a_0, ..., x_T, a_T)\}^{n}_{i=1}$ of agents videos from egocentric view and their associated body controls, such that every $x_j\in \mathbb{R}^{H\times W \times 3}$ is a video frame and $a_j\in\mathbb{R}^{d_{act}}$ an action in the Xsens skeleton ordering ~\citep{movella_mvn} for the upper body (everything above the pelvis), representing the change in translation, together with the delta rotation of all joints relative to the previous joint rotation. We represent motion in $3$D space, thus we have $3$ degrees of freedom for root translation, $15$ joints for the upper body and represent relative joint rotations as Euler angles in $3$D space leaving $d_{act} = 3 + 15 \times 3 = 48$.

We start by encoding each individual frame $s_i = \text{enc}(x_i)$ into a corresponding state representation, through a pre-trained VAE encoder~\citep{rombach2022high}. Given a sequence of controls $a_0, \dots a_T$, our goal is to build a generative model that captures the dynamics of the environment:
\begin{equation}
P(s_T, \dots s_0 | a_T, \dots a_0 ) = P(s_0)\prod^{T-1}_{t=0} P(s_{t+1} | s_{t}, \dots, s_0, a_T, \dots a_0 )
\end{equation}
To simplify the model, we factorize the distribution and make a Markov assumption that the next state is dependent on the last $k$ states and a single past action:
\begin{equation}
P(s_{t+1} | s_{t}, \dots, s_0, a_T, \dots a_0) = P(s_{t+1} | s_{t}, \dots, s_{t-k+1}, a_{t-1})
\end{equation}
We aim to train a model parametrized by $\theta$ that minimizes the negative log-likelihood: 
$$\hat{\theta} = \arg \min_{\theta}\left[ -\log P_\theta(s_0) -\sum^{T-1}_{t=0}\log P_{\theta}(s_{t+1} | s_{t}, \dots, s_{t-k+1}, a_t)\right]$$ 
We model each transition $P_{\theta}(s_{t+1} | s_{t}, \dots, s_{t-k+1}, a_t)$ using a Denoising Diffusion Probabilistic Model (DDPM)~\citep{ho2020denoising}, which maximizes the (reweighted) evidence lower bound (ELBO) of the log-likelihood.  For each transition, we define the forward diffusion process $
q(z_{\tau} \mid s_{t+1}) = \mathcal{N}(z_\tau; \sqrt{\bar{\alpha}_\tau} s_{t+1}, (1 - \bar{\alpha}_\tau)\mathbf{I})$, where $z_\tau$ is the noisy version of $s_{t+1}$ at noise timestep $\tau$, and $\bar{\alpha}_\tau$ is the cumulative product of noise scales. The reverse process is learned by training a neural network $\epsilon_\theta$ to predict the noise given $z_\tau$ and the conditioning context $c_t = (s_t, \dots, s_{t-k+1}, a_t)$. 

Then denoising loss term for a transition is given by:
\begin{equation}
\mathcal{L}_\text{simple, t} = \mathbb{E}_{\tau, \epsilon \sim \mathcal{N}(0, I)} \left[ \left\| \epsilon - \epsilon_\theta\left(\sqrt{\bar{\alpha}_\tau} s_{t+1} + \sqrt{1 - \bar{\alpha}_\tau} \epsilon, \; c_t, \; \tau \right) \right\|^2 \right]
\end{equation}
Where $\mathcal{L}_\text{simple, 0}$ is the loss term corresponding to the unconditional generation of $s_0$. Additionally, we also predict the covariances of the noise, and supervise them using the full variational lower bound loss $\mathcal{L}_{\text{vlb,t}}$ as proposed by~\citep{pmlr-v139-nichol21a}. 

Hence the final objective yields a (weighted) version of the ELBO for each term in the sequence:
\begin{equation}
\label{eq:loss}
\mathcal{L} = \sum^{T-1}_{t=0} \mathcal{L}_{simple, t} + \lambda \mathcal{L}_{vlb, t}
\end{equation}

Despite not being a lower bound of the log-likelihood, the reweighted ELBO works well in practice for image generation with transformers~\citep{pmlr-v139-nichol21a,Peebles_2023_ICCV}.

The advantage of our formulation is that it allows training in parallelized fashion using causal masking. Given a sequence of frames and actions, we can train on every prefix of the sequence in a single forward-backward pass. Next, we elaborate on the architecture of our model.

\subsection{Autoregressive Conditional Diffusion Transformer}
\label{sec:arch}
While prior work in navigation world models~\citep{bar2024navigation} focuses on simple control signals like velocity and heading, modeling whole-body human motion presacents significantly greater challenges. Human activities involve complex, coordinated movements across multiple degrees of freedom, with actions that are both temporally extended and physically constrained. This complexity necessitates architectural innovations beyond standard CDiT approaches.

To address these challenges, we extend the Conditional Diffusion Transformer (CDiT) architecture with several key modifications that enable effective modeling of whole-body motion:

\noindent \textbf{Random Timeskips.} Human activities often span long time horizons with actions that can take several seconds to complete. At the same time, videos are a raw signal which requires vast amounts of compute to process. To handle video more efficiently, we introduce random timeskips during training~(see Figure~\ref{fig:arch}a), and include the timeskip as an action to inform the model's prediction. This allows the model to learn both short-term motion dynamics and longer-term activity patterns. Learning long-term dynamics is particularly important for modeling activities like reaching, bending, or walking, where the full motion unfolds over multiple seconds. In practice, we sample $16$ video frames from a $32$ second window.

\noindent \textbf{Sequence-Level Training.} Unlike NWM which predicts single frames, we model the entire sequence of motion by applying the loss over each prefix of frames following Eq.~\ref{eq:loss}. We include an example of this in Figure~\ref{fig:arch}b. This is crucial because human activities exhibit strong temporal dependencies - the way someone moves their arm depends on their previous posture and motion. We enable efficient training by parallelizing across sequence prefixes through spatial-only attention in the current frame and past-frame-only attention for historical context (Figure~\ref{fig:arch}c). In practice we train models with sequences of 16 frames.

\noindent \textbf{Action Embeddings.} The high-dimensional nature of whole-body motion (joint positions, rotations, velocities) requires careful handling of the action space. We take the most simple strategy: we concatenate all actions in time $t$ into a $1D$ tensor which is fed to each AdaLN layer for conditioning~(see Figure~\ref{fig:arch}c).

These architectural innovations are essential for modeling the rich dynamics of human motion. By training on sequence prefixes and incorporating timeskips, our model learns to generate temporally coherent motion sequences that respect both short-term dynamics and longer-term activity patterns. The specialized action embeddings further enable precise control over the full range of human movement, from subtle adjustments to complex coordinated actions.

\subsection{Inference and Planning with \ours}

\noindent \textbf{Sampling procedure at test time.}
Given a set of context frames $(x_t,...,x_{t-k+1})$, we encode these frames to get $(s_t,...,s_{t-k+1})$ and pass the encoded context as the clean tokens in Figure \ref{fig:arch}b and pass in randomly sampled noise as the last frame. We then follow the DPPM sampling process to denoise the last frame conditioning on our action. For faster inference time, we employ special attention masks where we change the mask in Figure \ref{fig:arch}c for within image attention to only be applied on the tokens of the last frame and change the mask for cross attention to context so that cross attention is only applied for the last frame.

\noindent \textbf{Autoregressive rollout strategy.}
To follow a set of actions we use an autoregressive rollout strategy. Given an initial set of context frames we $(x_{t}, ..., x_{t-k+1})$ we start by encoding each individual frame to get $(s_{t}, ..., s_{t-k+1})$ and add the current action to create the conditioning context $c_t = (s_t,...,s_{t-k+1}, a_t)$. We then sample from our model parameterized by $\theta$ to generate the next state:
$s_{t+1} = P_{\theta}(s_{t+1} | c_{t})$. We then discard the first encoding and append the generated $s_{t+1}$ and add the next action to produce the next context $c_{t+1} = (s_{t+1}, s_t,..., s_{t-k+1}, a_{t+1})$. We then repeat the process for our entire set of actions. Finally, to visualize the predictions, we decode the latent states to pixels using the VAE decoder~\cite{rombach2022high}.

\section{Experiments and Results}
\label{sec:exp}

\subsection{Experiment Setting}
\label{sec:exp:setting}

\noindent \textbf{Dataset.}
We use the Nymeria dataset~\citep{ma2024nymeria}, which contains synchronized egocentric video and full-body motion capture, recorded in diverse real-world settings using an XSens system~\citep{movella_mvn}. Each sequence includes RGB frames and 3D body poses in the XSens skeleton format, covering global translation and rotations of body joints. We sample body motions at 4 FPS. Videos are center-cropped and resized to 224$\times$224. We split the dataset 80/20 for training and evaluation, and report all metrics on the validation set.

\noindent \textbf{Training Details.}
We train variants of Conditional Diffusion Transformer (CDiT-S to CDiT-XXL, up to 32 layers) using a context window of 3–15 frames and predicting 64-frame trajectories. Models operate on 2$\times$2 patches and are conditioned on both pose and temporal embeddings. We use AdamW (lr=$8\mathrm{e}{-5}$, betas=(0.9, 0.95), grad clip=10.0) and batch size 512. Action inputs are normalized to $[-1, 1]$ for translation and $[-\pi, \pi]$ for rotation. All experiments use  Stable Diffusion VAE tokenizer and follow NWM’s hardware and evaluation setup. Metrics are averaged over 5 samples per sequence.

\subsection{Comparison with Baselines}
\begin{figure}[h]
    \centering
    \footnotesize
    \vspace{-0.1cm}
    \begin{minipage}[t!]{0.49\textwidth}
        \centering
        \vspace{3mm}
        \captionof{table}{\textbf{Baseline Perceptual Metrics.} Comparison of baselines on single-step prediction 2 seconds ahead. }
        \resizebox{\textwidth}{!}{
        
        \begin{tabular}{l|cc|c}
            \toprule
            Model & LPIPS $\downarrow$ & DreamSim $\downarrow$ & FID $\downarrow$ \\
            \midrule
            DF$^{*}$ & $0.352^{0.003}$ & $0.244^{0.003}$ & $73.052^{1.101}$ \\
            CDiT & $0.313^{0.001}$ & $0.202^{0.002}$ & $63.714^{0.491}$ \\
            \rowcolor{gray!15}\ours & $0.303^{0.001}$ & $0.193^{0.002}$ & $62.293^{0.671}$ \\
            \bottomrule
        \end{tabular}}
        
        \label{tab:baseline}
    \end{minipage} %
    \hfill
    \begin{minipage}[t!]{0.49\textwidth}
        \centering
        \resizebox{\textwidth}{!}{
            \includegraphics[width=\textwidth]{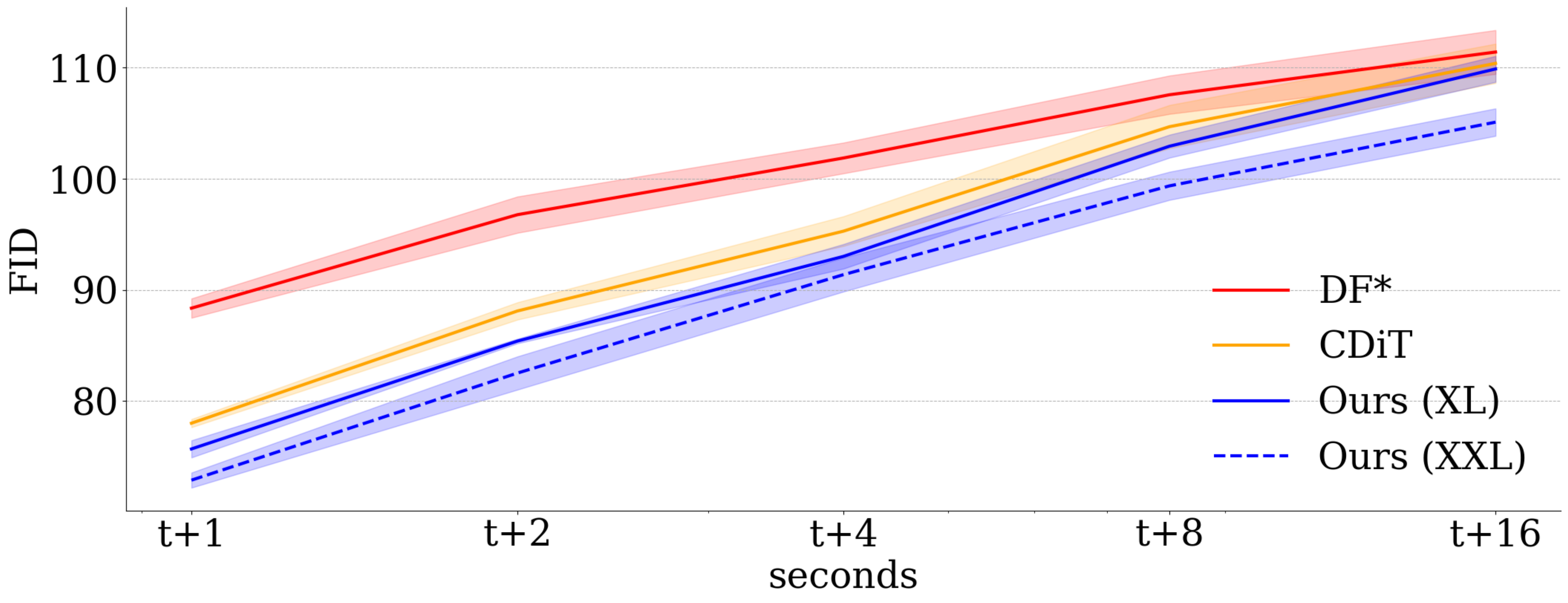}}
            
            \captionof{figure}{
            \textbf{Video Quality Across Time (FID).} Comparison of generation accuracy and quality as a function of time for up to $16$ seconds. Qualitative results for $16$ second rollouts can be seen in Figure~\ref{fig:teaser}c and Figure~\ref{fig:longh}. }
        \label{fig:video_quality}
        
    \end{minipage}
\end{figure}

\begin{figure}
    \centering
    \includegraphics[width=\linewidth]{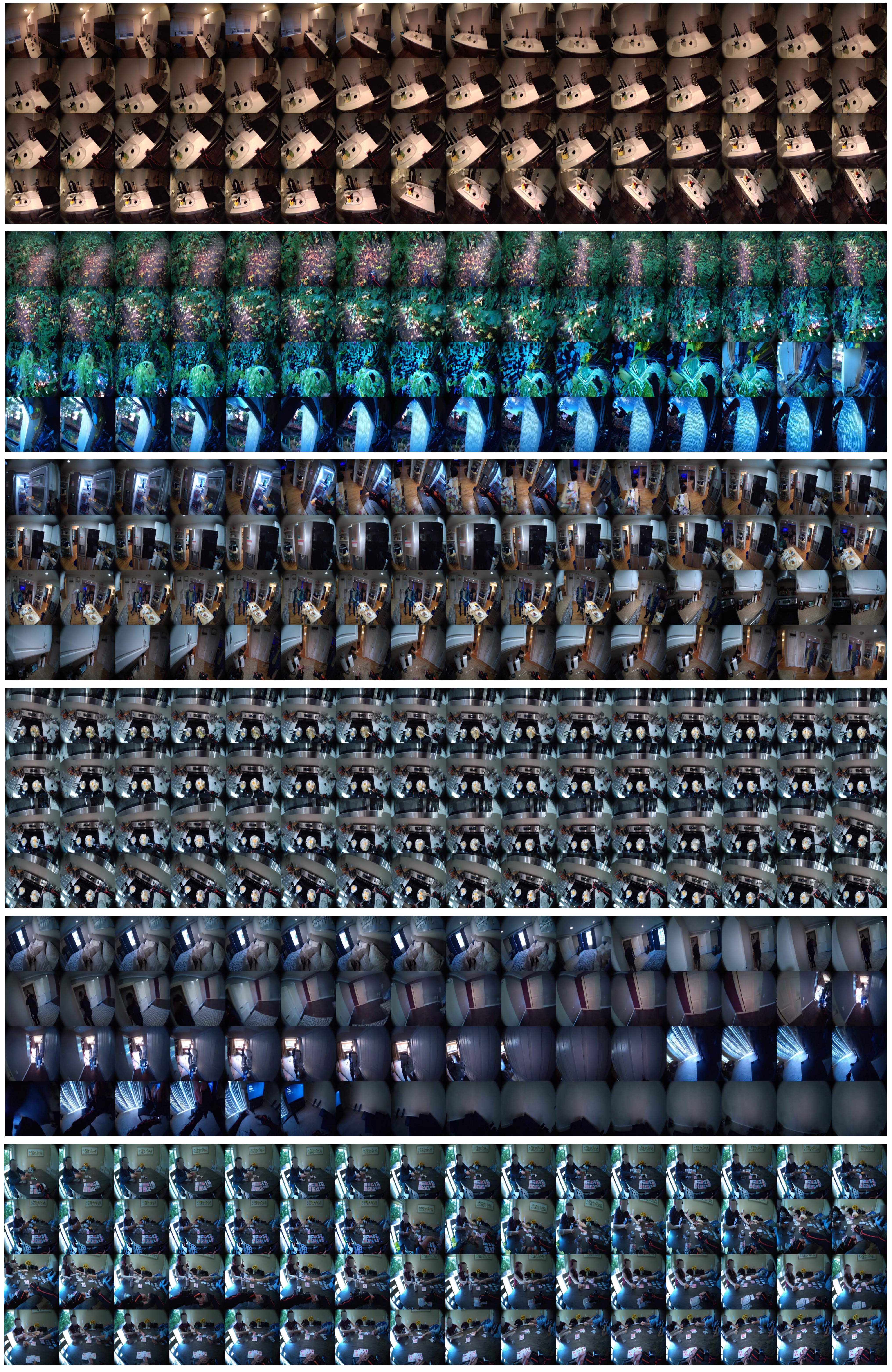}
    \caption{\textbf{Generation Over Long-Horizons}. We include $16$-second video generation examples. }
    \label{fig:longh}
\end{figure}

To comprehensively evaluate our model, we compare \ours{} with (CDiT~\cite{bar2024navigation} and Diffusion Forcing~\citep{chen2024diffusion}) along two key dimensions. First, to assess whether the model faithfully simulates future observations conditioned on actions, we use LPIPS~\citep{zhang2018unreasonable} and DreamSim~\citep{fu2023dreamsim}, which measure perceptual and semantic similarity to ground truth. Second, to evaluate the overall quality and realism of the generated samples, we report FID~\citep{heusel2017gans}. As shown in Table~\ref{tab:baseline}, our model achieves better results on both action consistency and generative quality. Furthermore, Figure~\ref{fig:video_quality} shows that our models tend to maintain lower FID scores than the baselines as the prediction horizon increases, suggesting improved visual quality and temporal consistency over longer rollouts. Qualitative results for $16$ second rollouts can be seen in Figure~\ref{fig:teaser}c and Figure~\ref{fig:longh}. We implement Diffusion Forcing (DF$^*$) on top of \ours~by applying the diffusion forward process to the entire sequence of encoded latents, then predicting the next state given the previous (noisy) latents. At test time, we autoregressively predict the next state as in \ours, without injecting noise into previously predicted frames, like~\cite{chen2024diffusion}.

\subsection{Atom Actions Control}
\begin{figure}
    \centering
    \includegraphics[width=\linewidth]{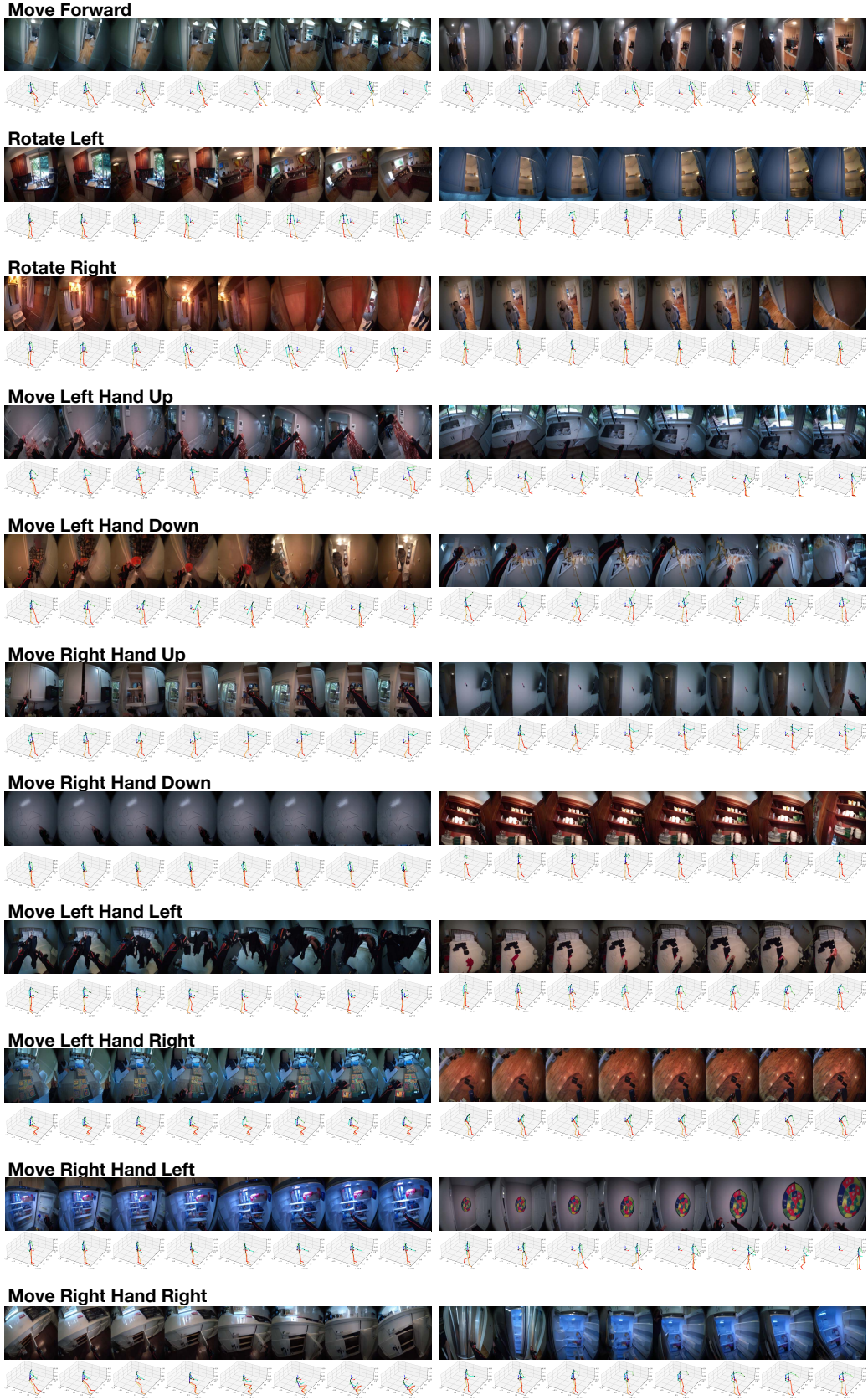}
    \caption{\textbf{Atom Actions Generation}. We include video generation examples of different atomic actions specified by 3D-body poses.}
    \label{fig:atom}
\end{figure}

\begin{figure}
    \centering
    \includegraphics[width=0.95\linewidth]{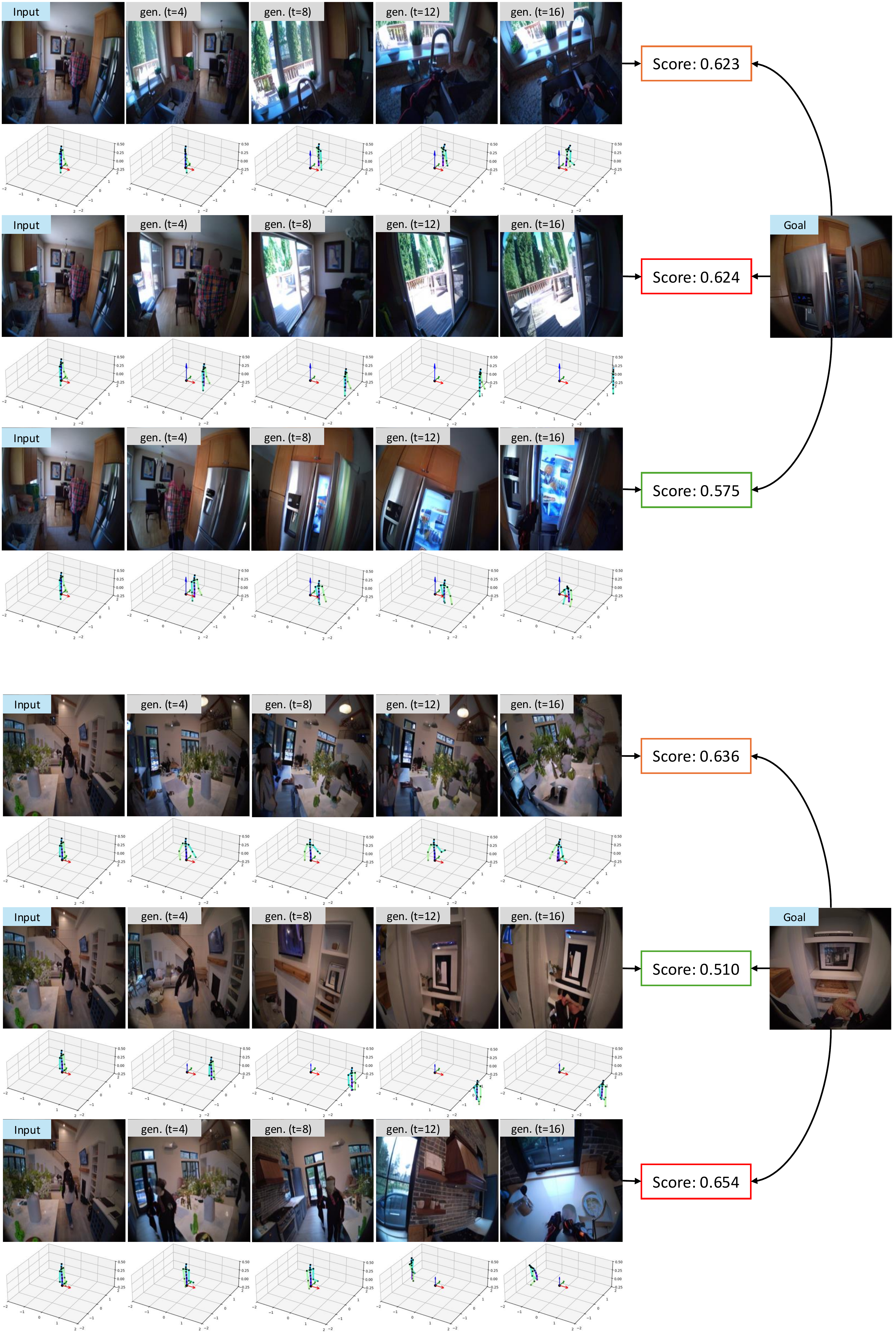}
    \caption{\textbf{Planning with Counterfactuals}. We demonstrate a planning example by simulating multiple action candidates using PEVA and scoring them based on their perceptual similarity to the goal, as measured by LPIPS~\citep{zhang2018unreasonable}. In the first case, we show that PEVA enables us to rule out action sequences that leads us to the sink in the top row, and outdoors in the second row. In the second case we show PEVA allows us to find a reasonable sequence of actions to open the refrigerator in the third row. \ours{} enables us to rule out action sequences that grab the nearby plants and go to the kitchen and mix ingredients. \ours{} allows us to choose the most correct action sequences that grab the box from the shelf. }
    \label{fig:counterfactuals}
\end{figure}

To better evaluate \ours{}’s ability to follow structured physical control, we decompose complex motions into atomic actions. By analyzing joint trajectories over short windows, we extract video segments exhibiting fundamental movements—such as hand motions (up, down, left, right) and whole-body actions (forward, rotate)—based on thresholded positional deltas. We sample 100 examples per action type to ensure balanced coverage, and evaluate single-step prediction 2 seconds ahead. Qualitative results are shown in Figure~\ref{fig:teaser}a and Figure~\ref{fig:atom}, and quantitative results in Table~\ref{tab:motion_control}.

\begin{table}[h]
    \centering
    \tiny
    \vspace{-0.5cm}
    \caption{\textbf{Atomic Action Performance.} Comparison of models in generating videos of atomic actions.}
    \setlength{\tabcolsep}{0.2pt}  
    \begin{tabular}{l|ccc|cccc|cccc}
        \toprule
        \multirow{2}{*}{Model} & \multicolumn{3}{c|}{Navigation} & \multicolumn{4}{c|}{Left Hand} & \multicolumn{4}{c}{Right Hand} \\
        \cmidrule{2-4} \cmidrule{5-8} \cmidrule{9-12}
        & Forward & Rot.L & Rot.R & Left & Right & Up & Down & Left & Right & Up & Down \\
        \midrule
        DF$^{*}$ & $0.393^{0.011}$ & $0.314^{0.006}$ & $0.279^{0.005}$ & $0.292^{0.009}$ & $0.306^{0.005}$ & $0.332^{0.008}$ & $0.323^{0.006}$ & $0.304^{0.006}$ & $0.315^{0.007}$ & $0.305^{0.005}$ & $0.296^{0.008}$ \\
        CDiT & $0.348^{0.004}$ & $0.284^{0.003}$ & $0.249^{0.004}$ & $0.258^{0.005}$ & $0.265^{0.009}$ & $0.279^{0.008}$ & $0.267^{0.004}$ & $0.286^{0.007}$ & $0.273^{0.004}$ & $0.277^{0.004}$ & $0.268^{0.002}$ \\
        Ours (XL) & $0.337^{0.006}$ & $0.277^{0.006}$ & $0.242^{0.007}$ & $0.244^{0.005}$ & $0.257^{0.004}$ & $0.272^{0.008}$ & $0.263^{0.003}$ & $0.271^{0.005}$ & $0.267^{0.003}$ & $0.268^{0.004}$ & $0.256^{0.009}$ \\
        \rowcolor{gray!15}Ours (XXL) & $0.325^{0.006}$ & $0.269^{0.005}$ & $0.234^{0.004}$ & $0.236^{0.003}$ & $0.241^{0.003}$ & $0.251^{0.004}$ & $0.247^{0.005}$ & $0.256^{0.007}$ & $0.254^{0.005}$ & $0.252^{0.004}$ & $0.245^{0.005}$ \\
        \bottomrule
    \end{tabular}
    \setlength{\tabcolsep}{6pt}  
    \label{tab:motion_control}
\end{table}

\subsection{Ablation Studies}
We conduct ablation studies to assess the impact of key design choices in PEVA, summarized in Table~\ref{tab:ablations}. First, increasing the context window from 3 to 15 frames consistently improves performance across all metrics, highlighting the importance of temporal context for egocentric prediction. Second, model scale plays a significant role: larger variants from \model-S to \model-XXL show steady gains in perceptual and semantic fidelity. Lastly, we compare two action embedding strategies—MLP-based encoding versus simple concatenation—and find that the latter performs competitively despite its simplicity, suggesting that our structured action representation already captures sufficient motion information. The gray-highlighted row denotes the default configuration in main experiments.

\begin{table}[!h]
    \centering
    \footnotesize
    \caption{\textbf{Model Ablations}. We evaluate the impact of different context lengths, action embedding methods, and model sizes on single-step prediction performance (2 seconds into the future).}
    \begin{tabular}{l|cccc}
        \toprule
        \multirow{2}{*}{Configuration} & \multicolumn{4}{c}{Metrics} \\
        \cmidrule{2-5}
        & LPIPS $\downarrow$ & DreamSim $\downarrow$ & PSNR $\uparrow$ & FID $\downarrow$ \\
        \midrule
        \multicolumn{5}{l}{\textit{Context Length}} \\
        \quad 3 frames & $0.304^{0.002}$ & $0.199^{0.003}$ & $16.469^{0.044}$ & $63.966^{0.421}$ \\ 
        \quad 7 frames & $0.304^{0.001}$ & $0.195^{0.002}$ & $16.443^{0.068}$ & $62.540^{0.314}$ \\ 
        \rowcolor{gray!15} \quad 15 frames & $0.303^{0.001}$ & $0.193^{0.002}$ & $16.511^{0.061}$ & $62.293^{0.671}$ \\
        \midrule
        \multicolumn{5}{l}{\textit{Action Representation}} \\
        \quad Action Embedding ($d=512$) & $0.317^{0.003}$ & $0.202^{0.002}$ & $16.195^{0.081}$ & $63.101^{0.341}$ \\ 
        \rowcolor{gray!15}\quad Action Concatenation & $0.303^{0.001}$ & $0.193^{0.002}$ & $16.511^{0.061}$ & $62.293^{0.671}$ \\
        \midrule
        \multicolumn{5}{l}{\textit{Model Size}} \\
        \quad \model-S & $0.370^{0.002}$ & $0.327^{0.002}$ & $15.743^{0.060}$ & $101.38^{0.450}$ \\
        \quad \model-B & $0.337^{0.001}$ & $0.246^{0.002}$ & $16.013^{0.091}$ & $74.338^{1.057}$ \\
        \quad \model-L & $0.308^{0.002}$ & $0.202^{0.001}$ & $16.417^{0.037}$ & $64.402^{0.496}$ \\
        \quad \model-XL & $0.303^{0.001}$ & $0.193^{0.002}$ & $16.511^{0.061}$ & $62.293^{0.671}$ \\
        \rowcolor{gray!15} \quad \model-XXL & $0.298^{0.002}$ & $0.186^{0.003}$ & $16.556^{0.060}$ & $61.100^{0.517}$ \\
        \bottomrule
    \end{tabular}
    \label{tab:ablations}
\end{table}

\subsection{Long-Term Prediction Quality}

We evaluate the model’s ability to maintain visual and semantic consistency over extended prediction horizons. As shown in Figure~\ref{fig:longh}, PEVA generates coherent 16-second rollouts conditioned on full-body motion. Table~\ref{fig:video_quality} reports DreamSim scores at increasing time steps, showing a gradual degradation from 0.178 (1s) to 0.390 (16s), indicating that predictions remain semantically plausible even far into the future.

\subsection{Planning with Multiple Action Candidates.}

We demonstrate a sample in which PEVA enables planning with multiple action candidates in Figure~\ref{fig:teaser}b and Figure~\ref{fig:counterfactuals}. We start by sampling multiple action candidates and simulate each action candidate using PEVA via autoregressive rollout. Finally, we rank each action candidate's final prediction by measuring LPIPS similarity with the goal image. We find that PEVA is effective in enabling planning through simulating action candidates.

\section{Failure Cases, Limitations and Future Directions}
\label{sec:fail}

While our model demonstrates promising results in predicting egocentric video from whole-body motion, several limitations remain that suggest directions for future work. First, our planning evaluation is preliminary—we only explore a simulation-based selection over candidate actions for only the left or right arm. While this provides an early indication that the model can anticipate visual consequences of body movement, it does not yet support long-horizon planning or full trajectory optimization. Extending PEVA to closed-loop control or interactive environments is a key next step. Second, the model currently lacks explicit conditioning on task intent or semantic goals. Our evaluation uses image similarity as a proxy objective. Future work could explore combining PEVA with high-level goal conditioning or integrating object-centric representations.

\subsection{Some planning attempts with \ours}
\label{sec:plan}
Here we describe how to use a trained \ours{} to plan action sequences to achieve a visual target. We formulate planning as an energy minimization problem and perform standalone planning in the same way as NWM~\citep{bar2024navigation} using the Cross-Entropy Method (CEM)~\citep{rubinstein1997optimization} besides minor modifications in the representation and initialization of the action. 

For simplicity, we conduct two experiments where we only predict moving either the left or right arm controlled by predicting the relative joint rotations represented as euler angles. For each respective arm we control only the shoulder, upper arm, forearm, and hand leaving our actions space as $4 \times 3 = 12$ where we have $(\Delta\phi_{\text{shoulder}}, \Delta\theta_{\text{shoulder}}, \Delta\psi_{\text{shoulder}},...,\Delta\phi_{\text{forearm}}, \Delta\theta_{\text{forearm}}, \Delta\psi_{\text{forearm}})$. We initialize mean  $(\mu_{\Delta\phi_{\text{shoulder}}}, \mu_{\Delta\theta_{\text{shoulder}}}, \mu_{\Delta\psi_{\text{shoulder}}},...,\mu_{\Delta\phi_{\text{forearm}}}, \mu_{\Delta\theta_{\text{forearm}}}, \mu_{\Delta\psi_{\text{forearm}}})$ and variance $(\sigma^2_{\Delta\phi_{\text{shoulder}}}, \sigma^2_{\Delta\theta_{\text{shoulder}}}, \sigma^2_{\Delta\psi_{\text{shoulder}}},...,\sigma^2_{\Delta\phi_{\text{forearm}}}, \sigma^2_{\Delta\theta_{\text{forearm}}}, \sigma^2_{\Delta\psi_{\text{forearm}}})$ as the mean and variance of the next action across the training dataset for these segments. 

\begin{table}[h]
\centering
\footnotesize
\caption{Mean and Variance of relative rotation as euler angles $(\phi, \theta, \psi)$ for arm segments computed across the training dataset. }
\begin{tabular}{llcc}
\toprule
\textbf{Segment} & \textbf{Statistic} & \textbf{Right Arm} & \textbf{Left Arm} \\
\midrule
\multirow{2}{*}{Shoulder} 
  & Mean     & $(0.0027,\ -0.0012,\ -0.0015)$ & $(0.0624,\ 0.0687,\ 0.1494)$ \\
  & Variance & $(0.0010,\ -0.0006,\ 0.0003)$   & $(0.0625,\ 0.0697,\ 0.1496)$   \\
\midrule
\multirow{2}{*}{Upper Arm} 
  & Mean     & $(0.0107,\ -0.0011,\ -0.0020)$ & $(0.1119,\ 0.1647,\ 0.1791)$ \\
  & Variance & $(-0.0062,\ -0.0004,\ -0.0013)$   & $(0.0991,\ 0.1593,\ 0.1611)$    \\
\midrule
\multirow{2}{*}{Forearm} 
  & Mean     & $(0.0068,\ -0.0035,\ 0.0077)$  & $(0.1937,\ 0.2107,\ 0.2261)$ \\
  & Variance & $(-0.0036,\ -0.0063,\ 0.0002)$   & $(0.1791,\ 0.2012,\ 0.2186)$    \\
\midrule
\multirow{2}{*}{Hand} 
  & Mean     & $(0.0065,\ 0.0001,\ 0.004, )$   & $(0.2417,\ 0.229,\ 0.2631)$ \\
  & Variance & $(-0.0024,\ -0.0032,\ -0.0001)$   & $(0.2126,\ 0.2237,\ 0.2475)$    \\
\bottomrule
\end{tabular}
\label{tab:cem_init}
\end{table}

We assume the action is a straight continuous motion. Thus we repeat this action for our sequence length, in our case $T=8$ and optimize the delta actions. The time interval between steps is fixed at $k=0.25$ seconds. All other hyperparameters remain the same as in NWM~\citep{bar2024navigation}. 

\subsection{Qualitative Results}
Due to time constraints, we focus our investigation on arm movements—arguably the most complex among body actions. While this remains an open problem, we present preliminary results using \ours{} with CEM for standalone planning. This setting simplifies the high-dimensional control space while still capturing key challenges of full-body coordination.

\begin{figure}[!h]
    \centering
    \includegraphics[width=0.8\linewidth]{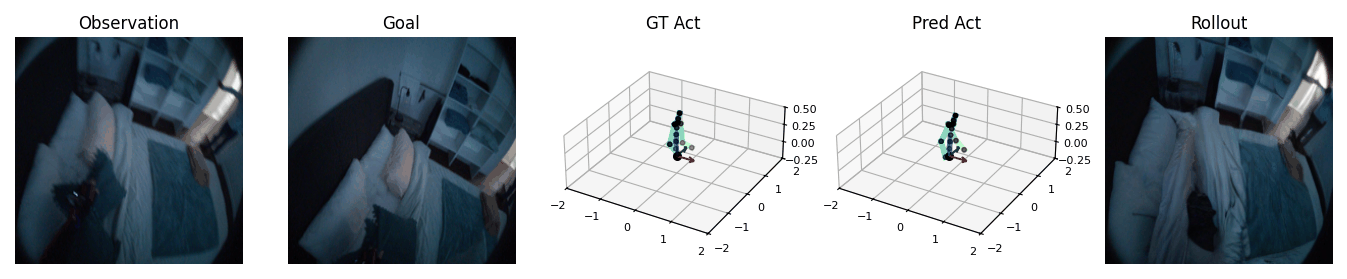}
    \caption{In this case, we are able to predict a sequence of actions that pulls our left arm in, similar to the goal. }
    \label{fig:cem_left_id_4}
\end{figure}

\begin{figure}[!h]
    \centering
    \includegraphics[width=0.8\linewidth]{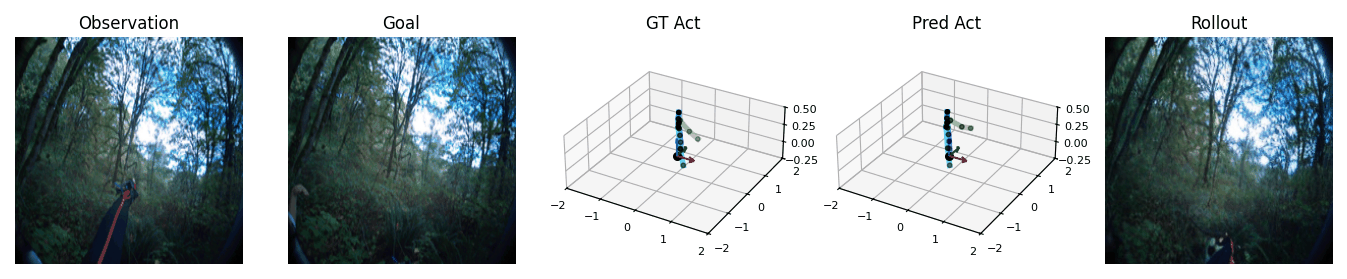}
    \caption{In this case, we are able to predict a sequence of actions that lowers our left arm, but not the same amount as the groundtruth sequence as we can see in the pose and hand at the bottom of our rollout. }
    \label{fig:cem_left_id_10}
\end{figure}

\begin{figure}[!h]
    \centering
    \includegraphics[width=0.8\linewidth]{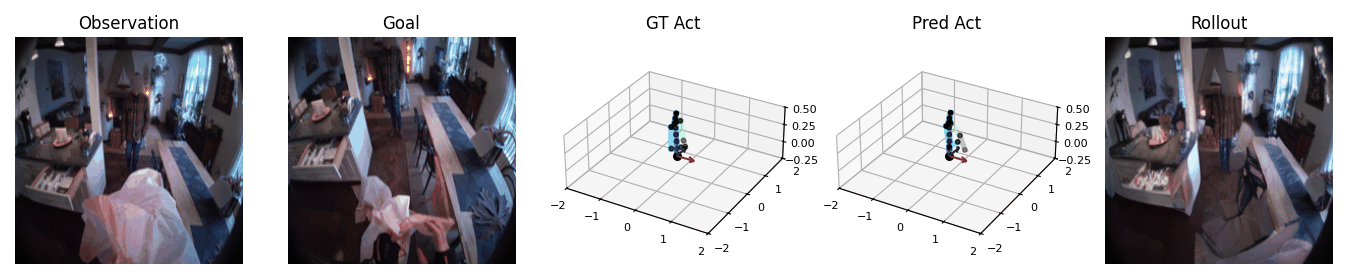}
    \caption{In this case, we are able to predict a sequence of actions that lowers our left arm that lowers the tissue. However, the goal image still has the tissue visible. }
    \label{fig:cem_left_id_15}
\end{figure}

\begin{figure}[!h]
    \centering
    \includegraphics[width=0.8\linewidth]{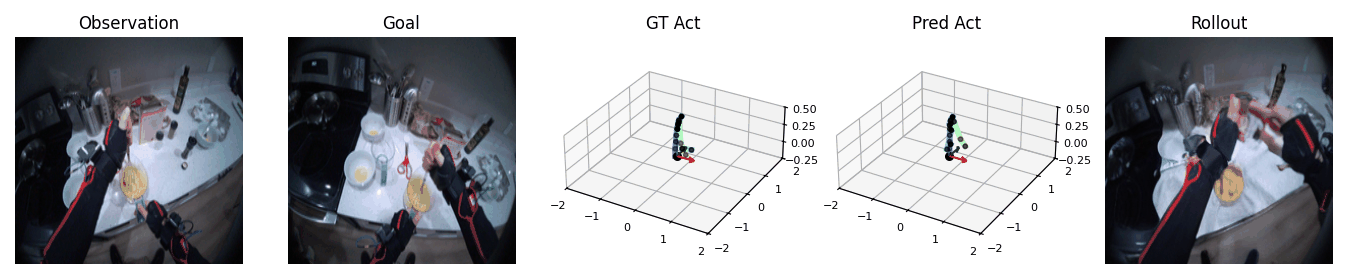}
    \caption{In this case, we are able to predict a sequence of actions that raises our right arm to the mixing stick. We see a limitation with our method as we only predict the right arm so we do not predict to move the left arm down accordingly. }
    \label{fig:cem_right_id_18}
    \vspace{-0.3cm}
\end{figure}

\begin{figure}[!h]
    \centering
    \includegraphics[width=0.8\linewidth]{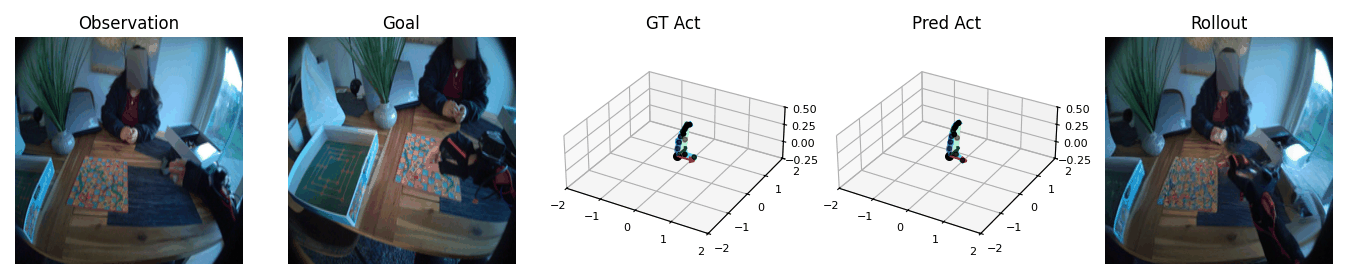}
    \caption{In this case, we are able to predict a sequence of actions that moves our right arm toward the left but not quite enough. We see a limitation with our method as we only predict the right arm so we do not predict any necessary additional body rotations. }
    \label{fig:cem_right_id_36}
\end{figure}

\begin{figure}[!h]
    \centering
    \includegraphics[width=0.8\linewidth]{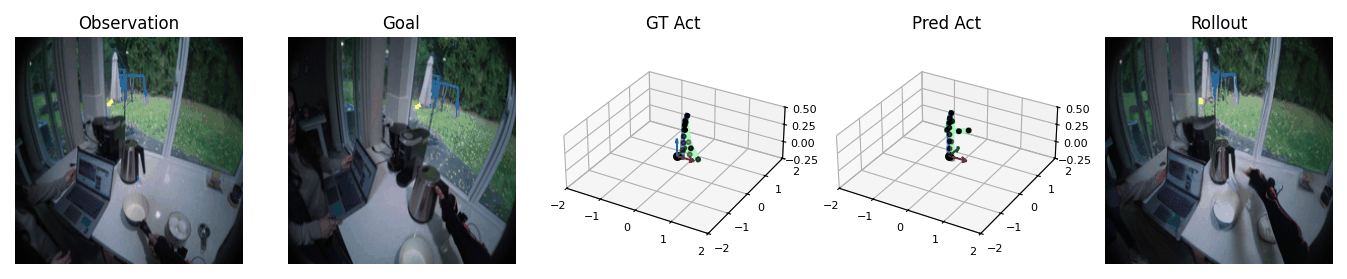}
    \caption{In this case, we are able to predict a sequence of actions that reaches toward the kettle but does not quite grab it as in the goal. }
    \label{fig:cem_right_kettle}
\end{figure}

\newpage
\section{Conclusion}
We introduced \model, a model that predicts egocentric video conditioned on detailed 3D human motion. Unlike prior work that uses low-dimensional or abstract control, \model{} leverages full-body pose sequences to simulate realistic and controllable visual outcomes. Built on a conditional diffusion transformer and trained on Nymeria, it captures the link between physical movement and first-person perception. Experiments show that PEVA improves prediction quality, semantic consistency, and fine-grained control over strong baselines. Our hierarchical evaluation highlights the value of whole-body conditioning across short-term, long-horizon, and atomic action tasks. While our planning results are preliminary, they demonstrate the potential for simulating action consequences in embodied settings. We hope this work moves toward more grounded models of perception and action for physically embodied intelligence.

\section*{Acknowledgment}
The authors thank Rithwik Nukala for his help in annotating atomic actions. We thank Katerina Fragkiadaki, Philipp Krähenbühl, Bharath Hariharan, Guanya Shi, Shubham Tsunami and Deva Ramanan for the useful suggestions and feedbacks for improving the paper; Jianbo Shi for the  discussion regarding control theory; Yilun Du for the support on Diffusion Forcing; Brent Yi for his help in human motion related works and Alexei Efros for the discussion and debates regarding world models.
This work is partially supported by the ONR MURI N00014-21-1-2801.

\newpage

{
    \small
    \bibliographystyle{ieeenat_fullname}
    \bibliography{main}
}

\medskip

\newpage

\section*{More Qualitative Results}
In the main paper, we provide three types of visualization: \ours{} can simulate counterfactuals, generate videos of atomic actions, and long video generation. 

Here, we show more qualitative results following the settings in main paper:

\begin{figure}[!h]
    \centering
    \includegraphics[width=0.8\linewidth]{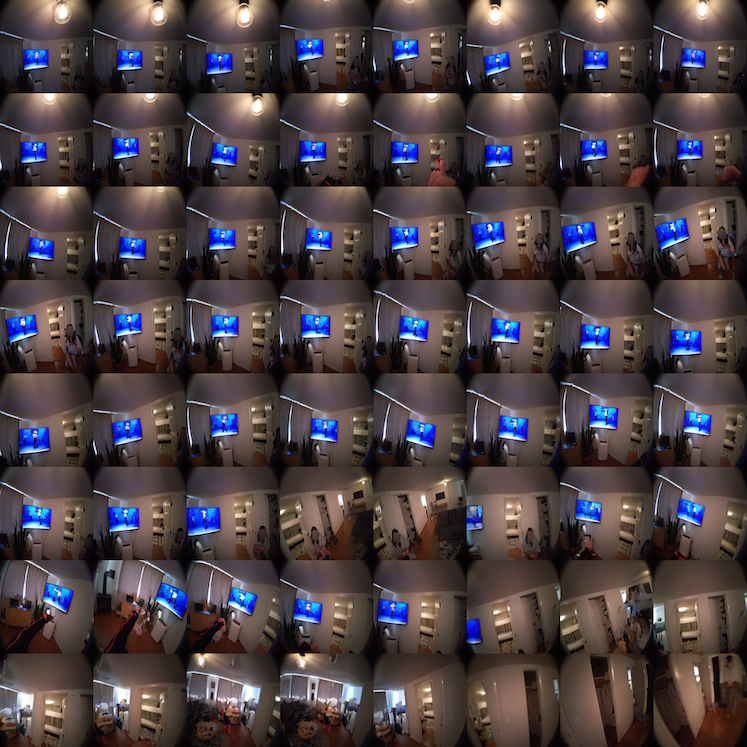}
    \caption{\textbf{Generation Over Long-Horizons}. We include $16$-second video generation examples.}
    \vspace{-0.2cm}
    \label{fig:sup_id_19}
    \vspace{-0.2cm}
\end{figure}

\begin{figure}
    \centering
    \includegraphics[width=0.8\linewidth]{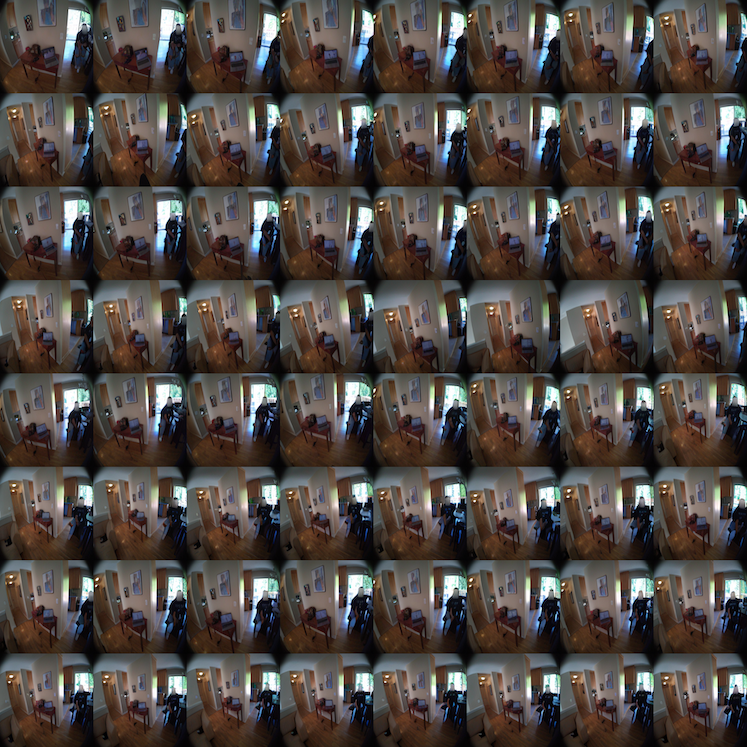}
    \vspace{-0.2cm}
    \label{fig:sup_id_22}
\end{figure}

\begin{figure}
    \centering
    \includegraphics[width=0.8\linewidth]{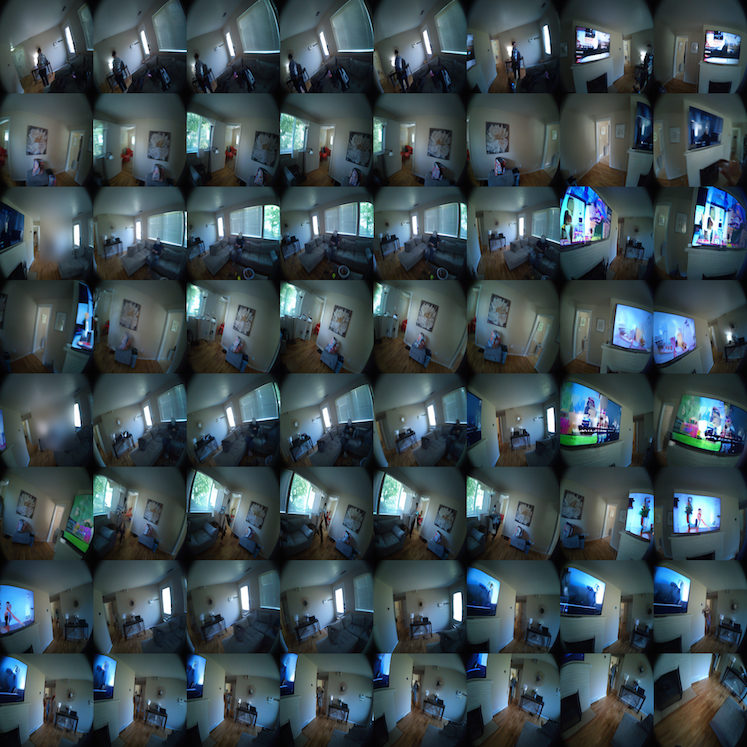}
    \caption{\textbf{Generation Over Long-Horizons}. We include $16$-second video generation examples.}
    \vspace{-0.2cm}
    \label{fig:sup_id_34}
    \vspace{-0.2cm}
\end{figure}

\begin{figure}
    \centering
    \includegraphics[width=0.8\linewidth]{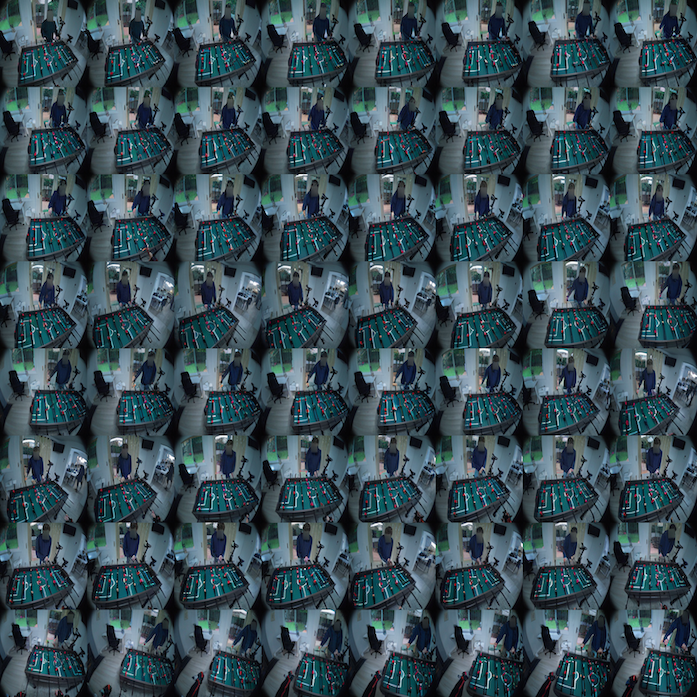}
    \vspace{-0.2cm}
    \label{fig:sup_id_40}
\end{figure}

\begin{figure}
    \centering
    \includegraphics[width=0.8\linewidth]{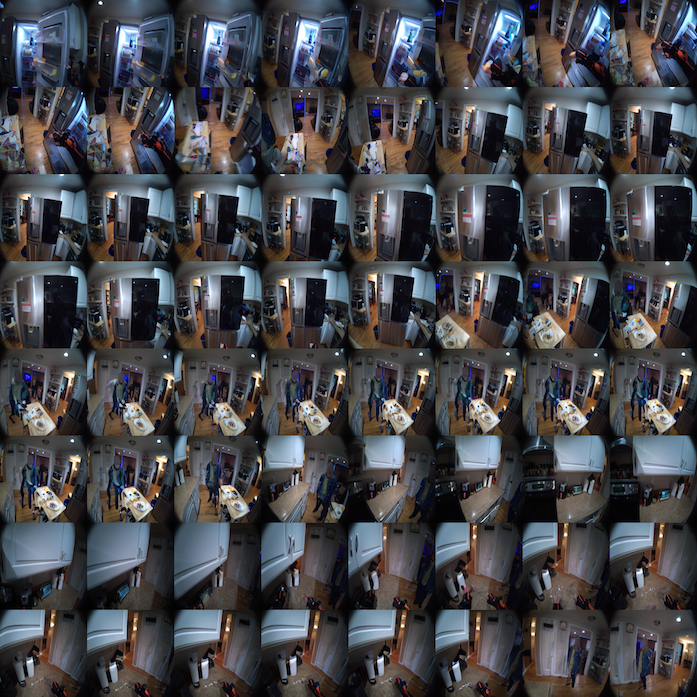}
    \caption{\textbf{Generation Over Long-Horizons}. We include $16$-second video generation examples.}
    \vspace{-0.2cm}
    \label{fig:sup_id_41}
    \vspace{-0.2cm}
\end{figure}

\begin{figure}
    \centering
    \includegraphics[width=0.8\linewidth]{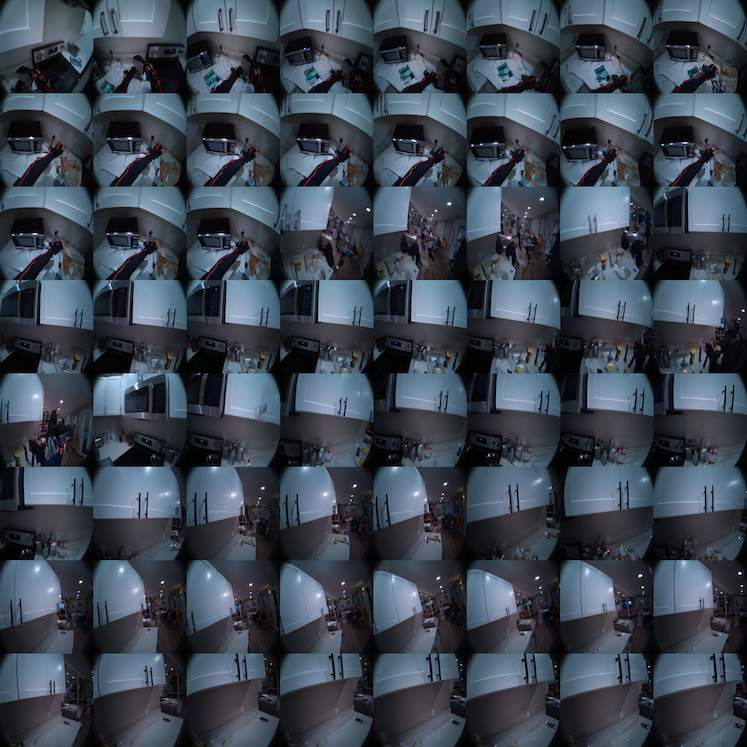}
    \vspace{-0.2cm}
    \label{fig:sup_id_44}
\end{figure}

\begin{figure}
    \centering
    \includegraphics[width=0.8\linewidth]{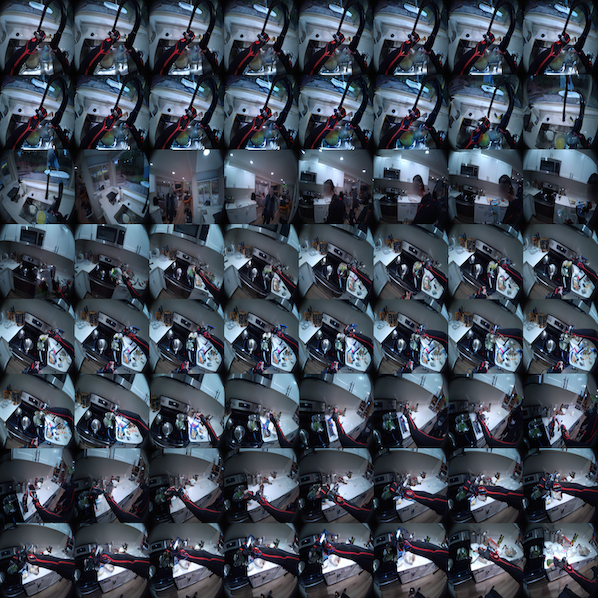}
    \caption{\textbf{Generation Over Long-Horizons}. We include $16$-second video generation examples.}
    \vspace{-0.2cm}
    \label{fig:sup_id_47}
    \vspace{-0.2cm}
\end{figure}

\begin{figure}
    \centering
    \includegraphics[width=0.8\linewidth]{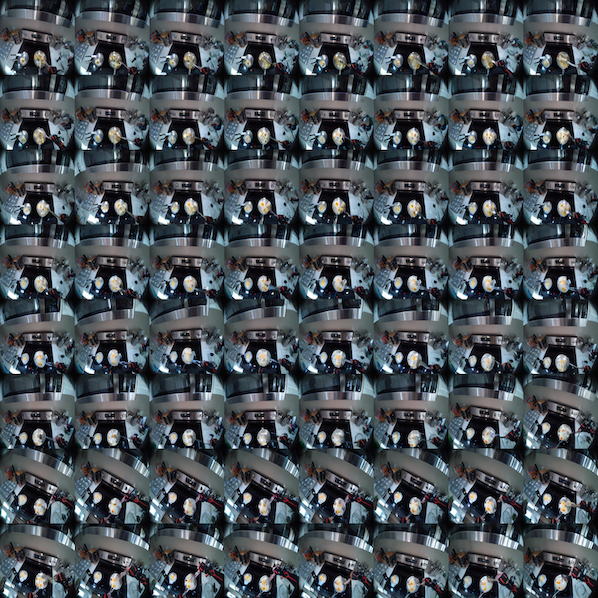}
    \vspace{-0.2cm}
    \label{fig:sup_id_53}
\end{figure}

\begin{figure}
    \centering
    \includegraphics[width=0.8\linewidth]{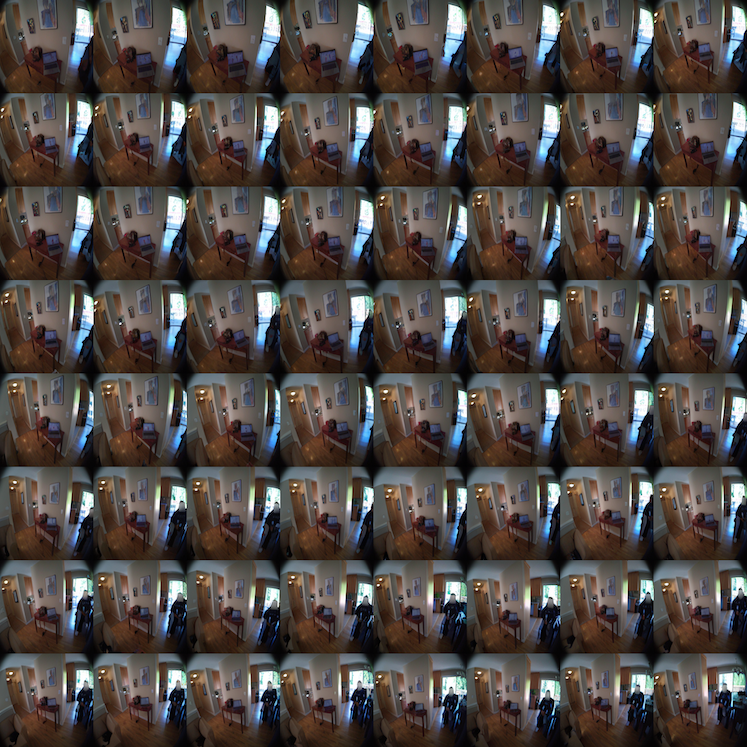}
    \caption{\textbf{Generation Over Long-Horizons}. We include $16$-second video generation examples.}
    \vspace{-0.2cm}
    \label{fig:sup_id_58}
    \vspace{-0.2cm}
\end{figure}

\begin{figure}
    \centering
    \includegraphics[width=0.8\linewidth]{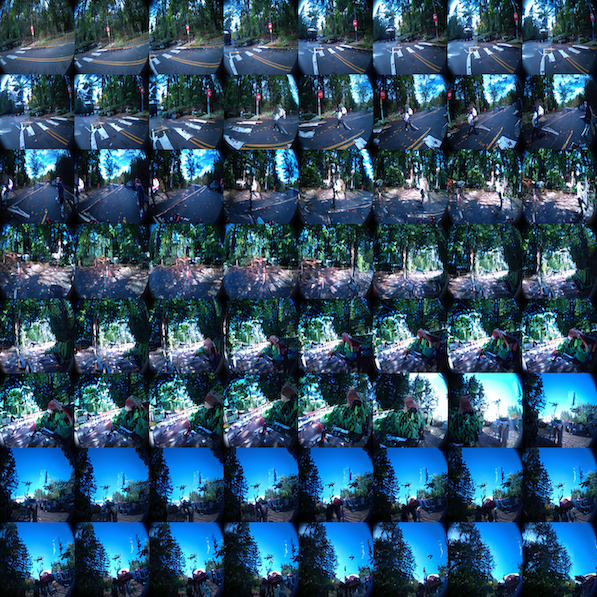}
    \vspace{-0.2cm}
    \label{fig:sup_id_65}
\end{figure}

\begin{figure}
    \centering
    \includegraphics[width=0.8\linewidth]{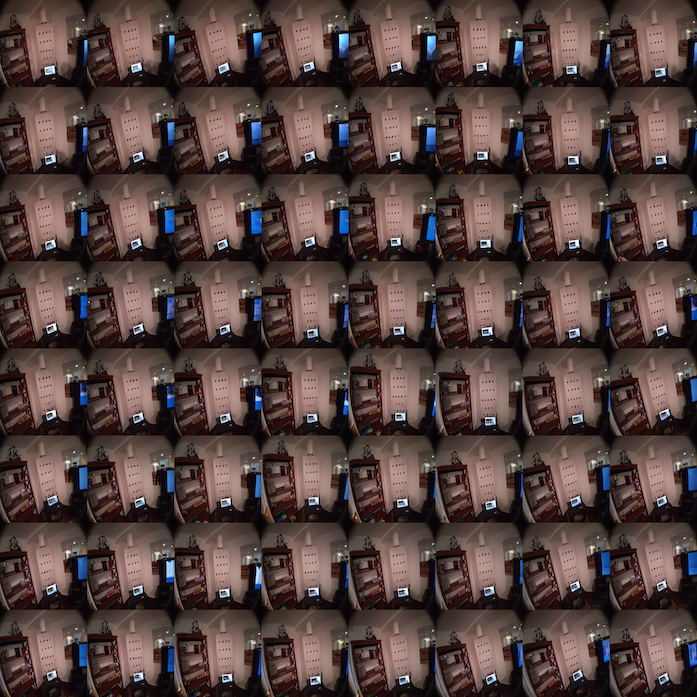}
    \caption{\textbf{Generation Over Long-Horizons}. We include $16$-second video generation examples.}
    \vspace{-0.2cm}
    \label{fig:sup_id_66}
    \vspace{-0.2cm}
\end{figure}

\begin{figure}
    \centering
    \includegraphics[width=0.8\linewidth]{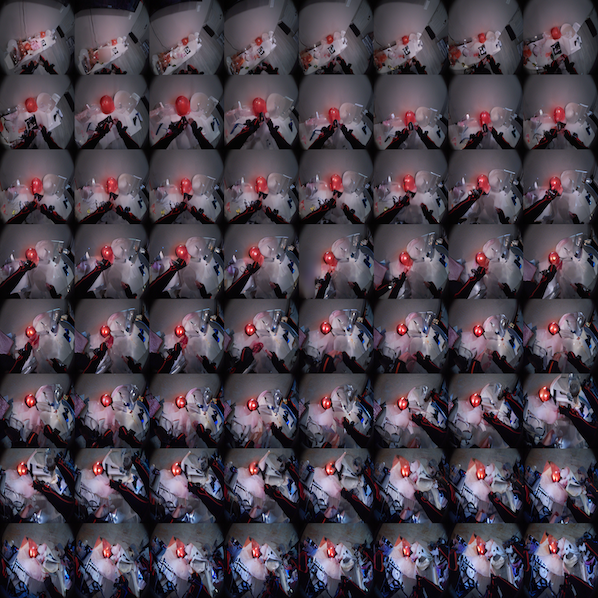}
    \vspace{-0.2cm}
    \label{fig:sup_id_67}
\end{figure}

\begin{figure}
    \centering
    \includegraphics[width=0.8\linewidth]{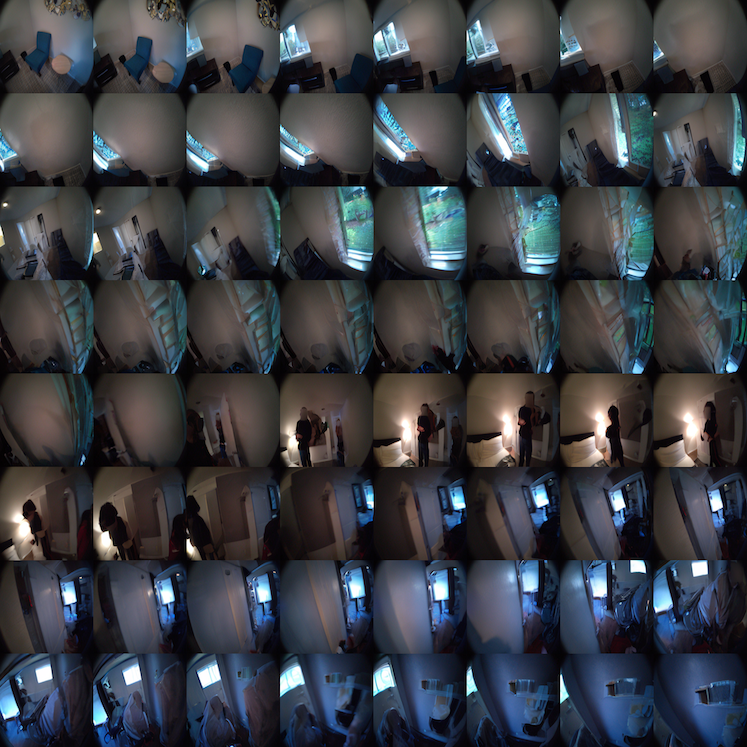}
    \caption{\textbf{Generation Over Long-Horizons}. We include $16$-second video generation examples.}
    \vspace{-0.2cm}
    \label{fig:sup_id_75}
    \vspace{-0.2cm}
\end{figure}

\begin{figure}
    \centering
    \includegraphics[width=0.8\linewidth]{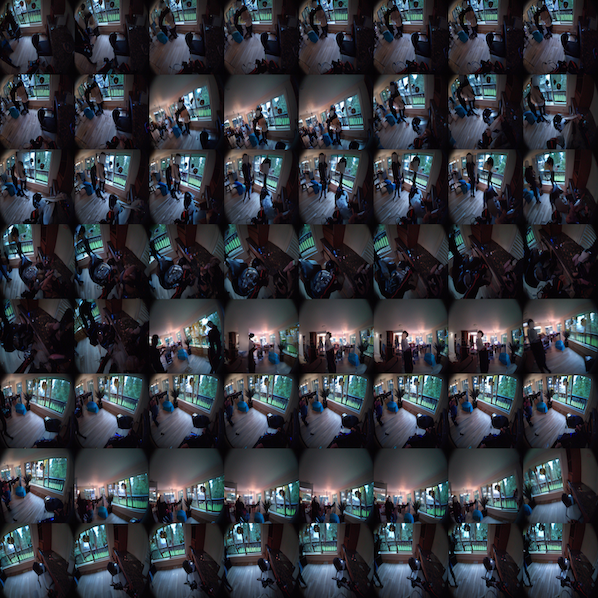}
    \vspace{-0.2cm}
    \label{fig:sup_id_83}
\end{figure}

\begin{figure}
    \centering
    \includegraphics[width=0.8\linewidth]{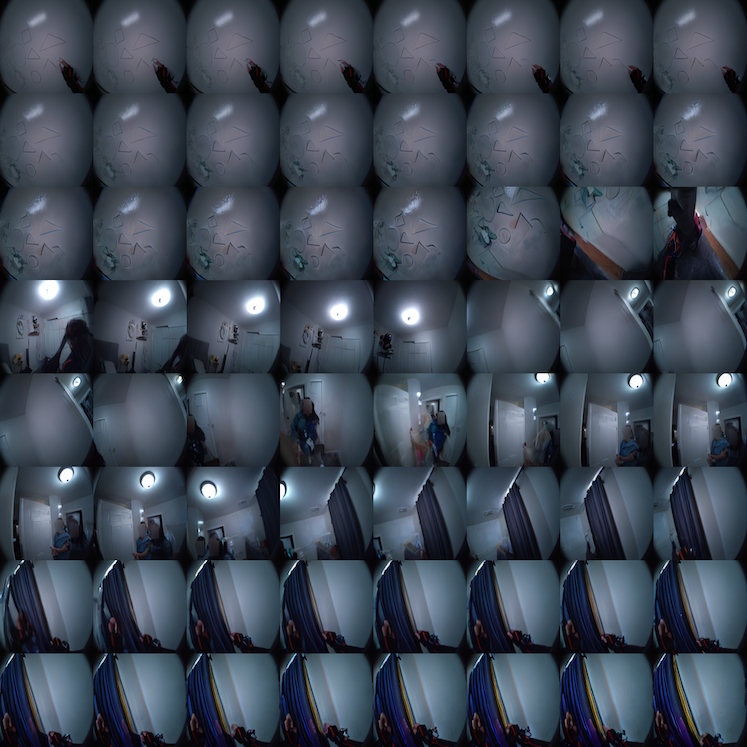}
    \caption{\textbf{Generation Over Long-Horizons}. We include $16$-second video generation examples.}
    \vspace{-0.2cm}
    \label{fig:sup_id_84}
    \vspace{-0.2cm}
\end{figure}

\begin{figure}
    \centering
    \includegraphics[width=0.8\linewidth]{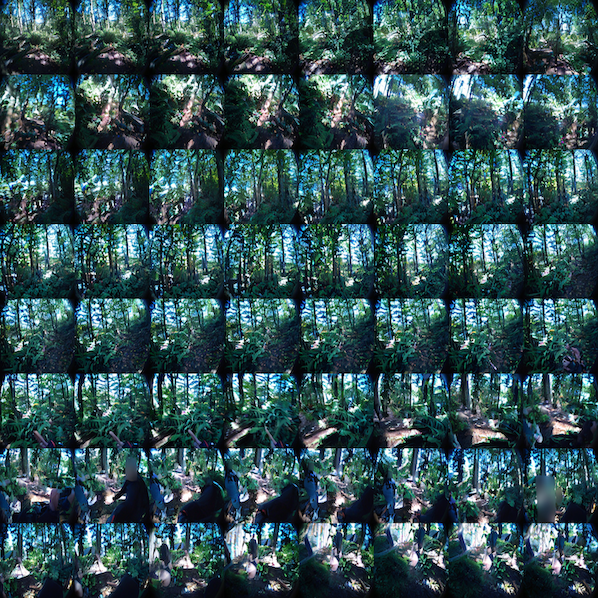}
    \vspace{-0.2cm}
    \label{fig:sup_id_86}
\end{figure}

\begin{figure}
    \centering
    \includegraphics[width=0.8\linewidth]{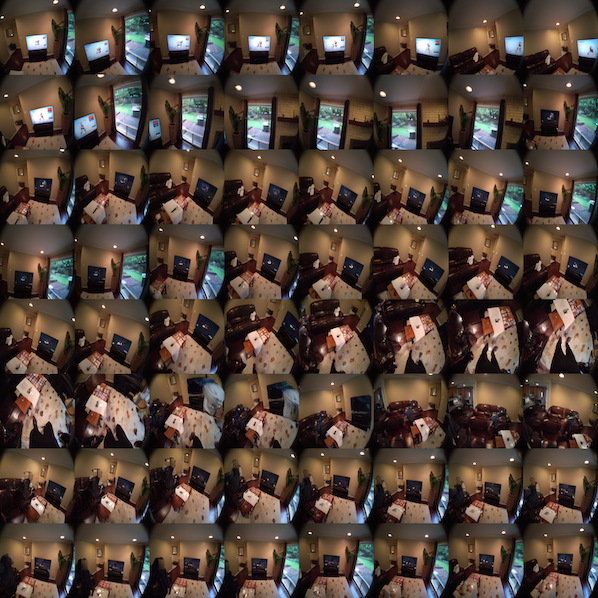}
    \caption{\textbf{Generation Over Long-Horizons}. We include $16$-second video generation examples.}
    \vspace{-0.2cm}
    \label{fig:sup_id_87}
    \vspace{-0.2cm}
\end{figure}

\begin{figure}
    \centering
    \includegraphics[width=0.8\linewidth]{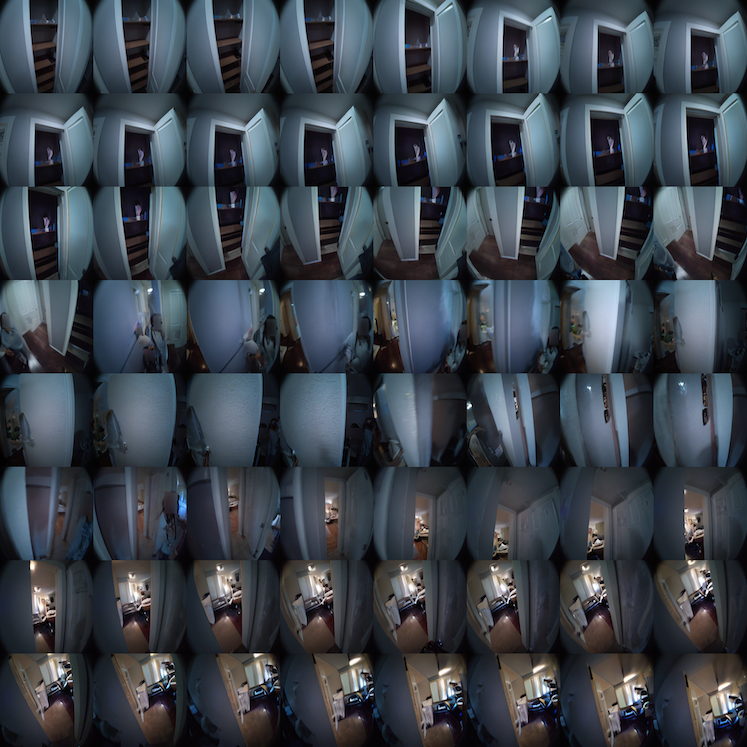}
    \vspace{-0.2cm}
    \label{fig:sup_id_88}
\end{figure}

\begin{figure}
    \centering
    \includegraphics[width=0.8\linewidth]{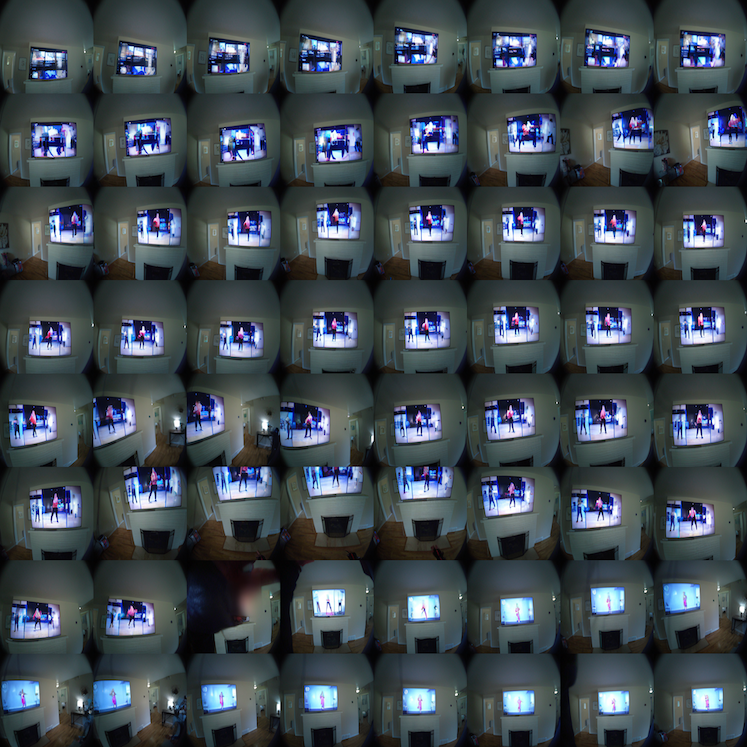}
    \caption{\textbf{Generation Over Long-Horizons}. We include $16$-second video generation examples.}
    \vspace{-0.2cm}
    \label{fig:sup_id_89}
    \vspace{-0.2cm}
\end{figure}

\begin{figure}
    \centering
    \includegraphics[width=0.8\linewidth]{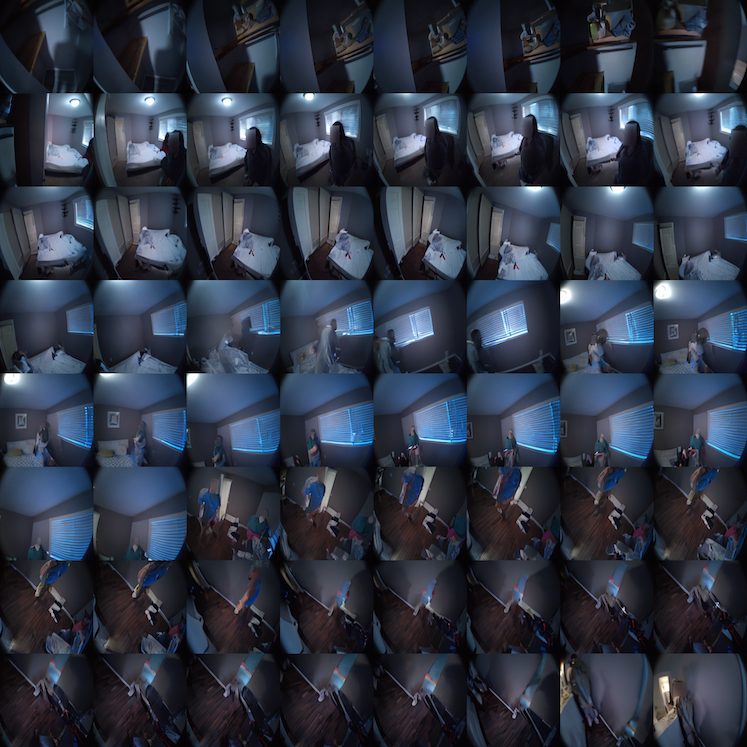}
    \vspace{-0.2cm}
    \label{fig:sup_id_90}
\end{figure}

\begin{figure}
    \centering
    \includegraphics[width=0.8\linewidth]{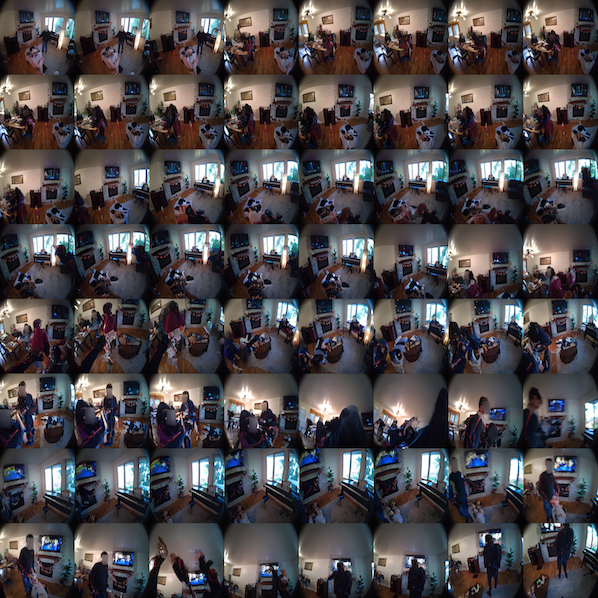}
    \caption{\textbf{Generation Over Long-Horizons}. We include $16$-second video generation examples.}
    \vspace{-0.2cm}
    \label{fig:sup_id_91}
    \vspace{-0.2cm}
\end{figure}

\begin{figure}
    \centering
    \includegraphics[width=0.8\linewidth]{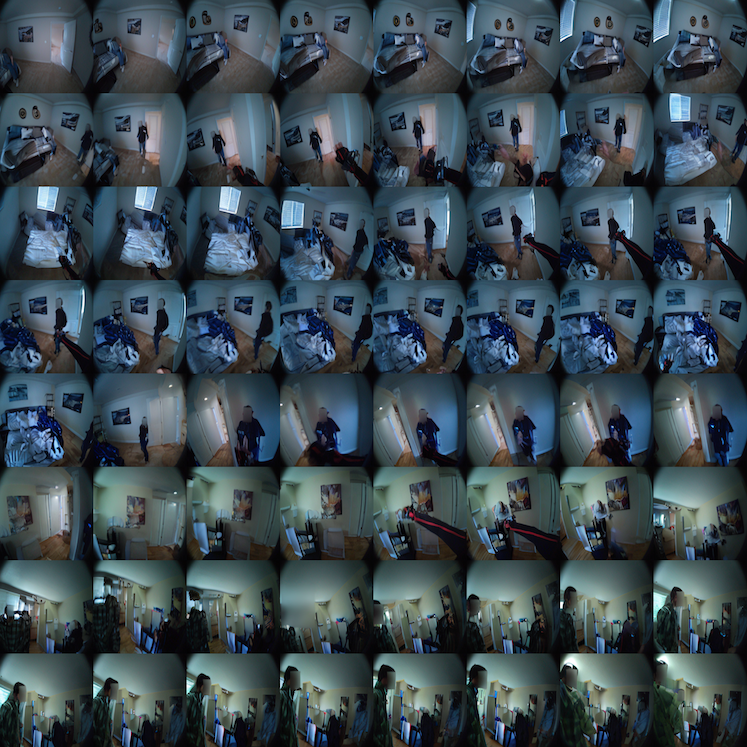}
    \vspace{-0.2cm}
    \label{fig:sup_id_92}
\end{figure}

\begin{figure}
    \centering
    \includegraphics[width=0.8\linewidth]{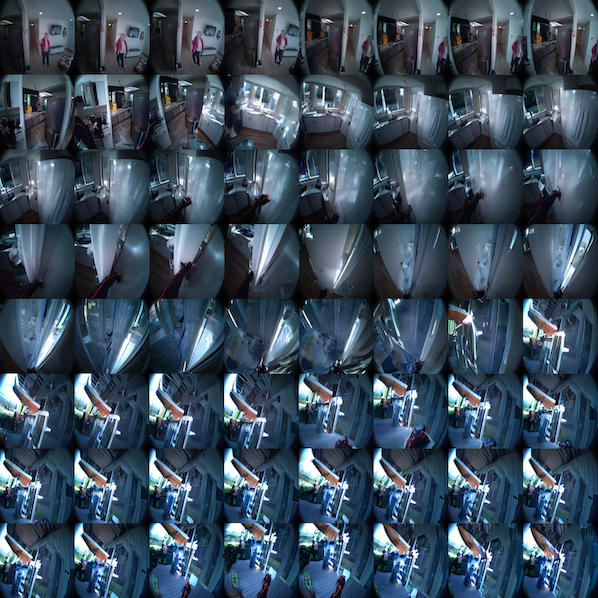}
    \caption{\textbf{Generation Over Long-Horizons}. We include $16$-second video generation examples.}
    \vspace{-0.2cm}
    \label{fig:sup_id_94}
    \vspace{-0.2cm}
\end{figure}

\begin{figure}
    \centering
    \includegraphics[width=0.8\linewidth]{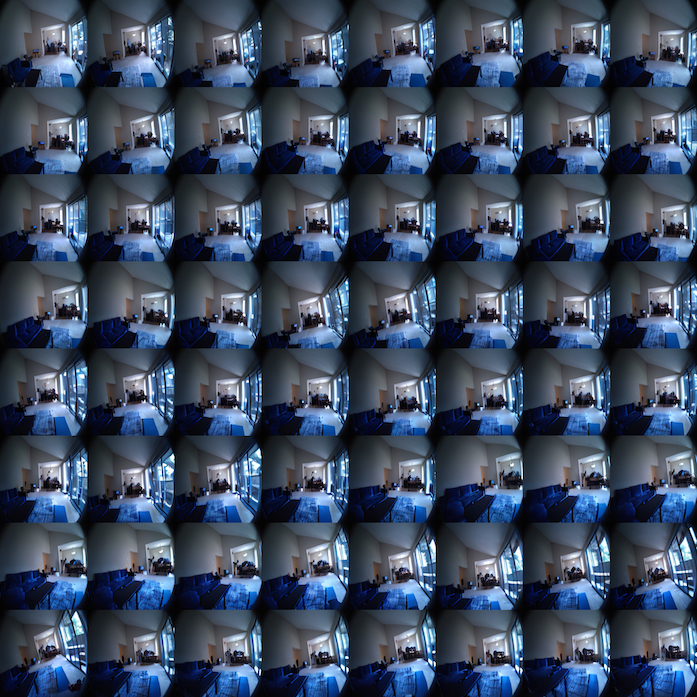}
    \vspace{-0.2cm}
    \label{fig:sup_id_104}
\end{figure}

\begin{figure}
    \centering
    \includegraphics[width=0.8\linewidth]{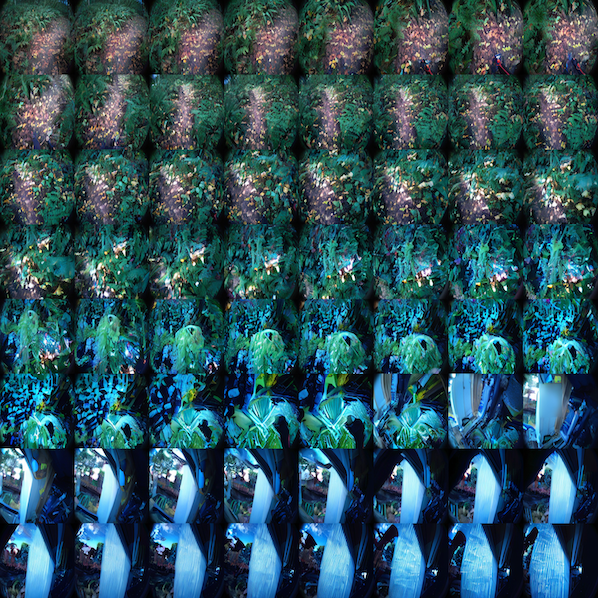}
    \caption{\textbf{Generation Over Long-Horizons}. We include $16$-second video generation examples.}
    \vspace{-0.2cm}
    \label{fig:sup_id_107}
    \vspace{-0.2cm}
\end{figure}

\medskip

\end{document}